\documentclass[11pt]{article}

\usepackage{fullpage}
\usepackage[ruled, linesnumbered, vlined, commentsnumbered]{algorithm2e}
\usepackage{graphicx,subfig} % figure related
\usepackage{amsfonts,amssymb,amsmath,amsthm,amsopn,mathtools}	% math related
\usepackage{booktabs,diagbox,colortbl,multirow,tabularx,threeparttable,hhline}
\usepackage[listings,skins,breakable]{tcolorbox}

\usepackage{enumerate}
\usepackage{authblk}
\usepackage{footnote}
\usepackage{hyperref}
\usepackage{prettyref}
\usepackage{cite}
\usepackage{setspace}
\usepackage{color}
\usepackage{xcolor}  % Required for custom colors
\usepackage{scrpage2}
\usepackage{geometry}

\usepackage{tikz}
\usepackage{pgfplots}
\usetikzlibrary{positioning,shapes,shadows,arrows,calc}
\tikzstyle{component}=[rectangle, draw=black, rounded corners, fill=blue!40, drop shadow, text centered, anchor=north, text=white, minimum height=1cm]
\tikzstyle{arrow}=[->, thick]

\pgfplotsset{compat=1.12}
\usetikzlibrary{intersections}
\usetikzlibrary{pgfplots.statistics}
\usepgfplotslibrary{fillbetween}

\definecolor{myblue}{RGB}{34,31,217}
\definecolor{mycyan}{gray}{.7}
\definecolor{Gray}{gray}{0.9}

\newtheorem{remark}{Remark}
\newtheorem{theorem}{Theorem}

\newtheorem{corollary}{Corollary}
\newtheorem{definition}{Definition}

\DeclareMathOperator*{\argmin}{argmin}

\newcommand{\pref}{\prettyref}

\newrefformat{fig}{Fig.~\ref{#1}}
\newrefformat{tab}{Table~\ref{#1}}
\newrefformat{sec}{Section~\ref{#1}}
\newrefformat{alg}{Algorithm~\ref{#1}}
\newrefformat{property}{Property~\ref{#1}}
\newrefformat{theorem}{Theorem~\ref{#1}}
\newrefformat{definition}{Definition~\ref{#1}}
\newrefformat{corollary}{Corollary~\ref{#1}}
\newrefformat{lemma}{Lemma~\ref{#1}}
\newrefformat{conj}{Conjecture~\ref{#1}}
\newrefformat{def}{Definition~\ref{#1}}
\newrefformat{eq}{equation~(\ref{#1})}
\newrefformat{app}{Appendix~\ref{#1}}

\usepackage{lscape}

\begin{document}

%% title
\title{\vspace{-1ex}\LARGE\textbf{Batched Data-Driven Evolutionary Multi-Objective Optimization Based on Manifold Interpolation}\footnote{Both authors made equal contributions to this paper.}~\footnote{This manuscript is submitted for potential publication. Reviewers can use this version in peer review.}}

%% authors and affiliations
\author[1]{\normalsize Ke Li}
\author[2]{\normalsize Renzhi Chen}
\affil[1]{\normalsize Department of Computer Science, University of Exeter, EX4 4QF, Exeter, UK}
\affil[2]{\normalsize PLA Academy of Military Science, Beijing, China}
\affil[$\ast$]{\normalsize Email: \texttt{k.li@exeter.ac.uk}}

\date{}
\maketitle

\vspace{-3ex}
{\normalsize\textbf{Abstract: } }Multi-objective optimization problems are ubiquitous in real-world science, engineering and design optimization problems. It is not uncommon that the objective functions are as a black box, the evaluation of which usually involve time-consuming and/or costly physical experiments. Data-driven evolutionary optimization can be used to search for a set of non-dominated trade-off solutions, where the expensive objective functions are approximated as a surrogate model. In this paper, we propose a framework for implementing batched data-driven evolutionary multi-objective optimization. It is so general that any off-the-shelf evolutionary multi-objective optimization algorithms can be applied in a plug-in manner. In particular, it has two unique components: 1) based on the Karush-Kuhn-Tucker conditions, a manifold interpolation approach that explores more diversified solutions with a convergence guarantee along the manifold of the approximated Pareto-optimal set; and 2) a batch recommendation approach that reduces the computational time of the optimization process by evaluating multiple samples at a time in parallel. Experiments on 136 benchmark test problem instances with irregular Pareto-optimal front shapes against six state-of-the-art surrogate-assisted EMO algorithms fully demonstrate the effectiveness and superiority of our proposed framework. In particular, our proposed framework is featured with a faster convergence and a stronger resilience to various PF shapes.


% !tex root = main.tex

\section{Introduction}
\label{sec:introduction}
Real-world problems in science, engineering and design often involve multiple conflicting objectives, as known as multi-objective optimization problems (MOPs). For example, in the optimal design of a water distribution system, many indicators need to be considered to achieve a trade-off between capital and/or operational cost and performance type benefits such as pressure deficit, reliable configurations under abnormal conditions and network resilience index. There does not exist a global optimal solution that optimizes all conflicting objectives. Instead, multi-objective optimization mainly aim to find a set of trade-off alternatives that compromise among different objectives before being handed over for multi-criterion decision-making.

Evolutionary algorithms (EAs) have been widely recognized as a major approach for multi-objective optimization given its population-based property for approximating a set of non-dominated solutions in a single run~\cite{Deb01}. Over the past three decades and beyond, many efforts have been dedicated to the developments of evolutionary multi-objective optimization (EMO) algorithms. According to their environmental selection mechanisms, the existing EMO algorithms can be classified into three categorizes: $1$) dominance-based methods, e.g., elitist non-dominated sorting genetic algorithm (NSGA-II)~\cite{DebAPM02}, $2$) indicator-based methods, indicator-based EA (IBEA)~\cite{ZitzlerK04}, and $3$) decomposition-based methods, e.g., multi-objective EA based on decomposition (MOEA/D)~\cite{ZhangL07,LiZ19}.

In practice, it is not uncommon that the objective functions of real-world problems are as a black box and are expensive to evaluate, either computationally or economically. For example, computational fluid dynamic simulations can take from minutes to hours to carry out a single function evaluation (FE)~\cite{JinS09}. Due to the population-based and iterative nature, EAs usually require a vast amount of FEs to obtain reasonably acceptable solutions. This is unrealistic when FEs are expensive thus it significantly hinders a wider uptake of EAs in real-world scenarios. To alleviate this issue, surrogate models, built by data collected from expensive FEs, have emerged as a powerful tool to assist EAs for solving expensive optimization problems, also known as data-driven evolutionary optimization\footnote{It is called surrogate-assisted EA interchangeably~\cite{Jin05} in the literature.}~\cite{JinWCGM19}. It consists of three intertwined design components.
\begin{itemize}
    \item The first one is the \underline{surrogate modeling} of the expensive objective functions. Many off-the-shelf machine learning approaches, e.g., support vector machine (SVM)~\cite{LoshchilovSS10}, Gaussian process regression (GPR) or Kriging model~\cite{Knowles06,ZhangLTV10,ChughJMHS18} and radial basis function networks (RBFN)~\cite{SunLGHL11,MartinezC13,AkhtarS16}, can serve this purpose. 
    \item The other one is the \underline{surrogate-assisted search process} either directly driven by the surrogate objective functions or the uncertainty inferred from the model, also known as the acquisition function, e.g., expected improvement~\cite{Mockus94}, upper confidence bound~\cite{SrinivasKKS10} and probability of improvement~\cite{Kushner64}, in GPR-assisted EAs~\cite{ShahriariSWAF16}. 
    \item The last one is the \underline{model management} that mainly aims to select promising solution(s) output from the search process for conducting expensive FEs. These newly evaluated solutions will thus be used to update the surrogate model accordingly.
\end{itemize}

In practice, many physical experiments can be carried out in parallel given the availability of more than one infrastructure. For example, laboratory technicians often perform experiments with duplicated setups in parallel to mitigate empirical bias. Likewise, in automated machine learning, training and validating machine learning models are usually distributed into multiple cores or GPUs for hyperparameter optimization. An effective parallelization, also known as batch recommendations/evaluations in data-driven evolutionary optimization, are of practical importance to significantly save the computational time by reducing the number of iterations. However, this line of research is relatively lukewarm in the data-driven evolutionary optimization community~\cite{ZhangLTV10,ZhanCL17,GuoJDC19}.

As discussed in~\cite{IshibuchiHS19}, it is unrealistic to have a regular Pareto-optimal front (PF) in real-world MOPs. On the contrary, due to the complex and non-linear relationship between objectives, it is not uncommon to have an irregular PF featured as disconnected, incomplete, degenerated, and/or badly-scaled. Although there have been growing interests for handling MOPs with irregular PFs in the EMO community (e.g.,~\cite{GuC18,WuLKZZ19,LiuIMN20}), it has rarely been considered in the context of data-driven EMO, except for~\cite{HabibSCRM19}.

To address the above issues, this paper proposes a batched data-driven EMO framework based on manifold interpolation for solving expensive MOPs with various PF shapes. It consists of the following four major design components.
\begin{itemize}
    \item\texttt{Surrogate modeling}: GPR is used to build the surrogate model for each computationally expensive objective function.
    \item\texttt{Evolutionary search}: This step searches for an approximated PF based on the surrogate objective functions. In particular, any existing EMO algorithm can be used to serve this purpose where we use NSGA-II, IBEA and MOEA/D for proof-of-concept purposes.
    \item\texttt{Manifold interpolation}: Based on the Karush-Kuhn-Tucker (KKT) conditions~\cite{KuhnT51}, this step is designed to interpolate new candidate solutions along the manifold of the approximated surrogate Pareto-optimal set with regard to the non-dominated solutions obtained in the \texttt{evolutionary search} step.
    \item\texttt{Batch recommendation}: Two types of simple and effective batch recommendation mechanisms are proposed to pick up multiple candidate solutions from the non-dominated solutions obtained in the \texttt{manifold interpolation} step for expensive FEs. In particular, one is directly derived from the native environmental selection mechanism of the EMO algorithm used in the \texttt{evolutionary search} step while the other is based on the individual Hypervolume contribution.
\end{itemize}
In our experiments, we generate six algorithm instantiations of our proposed framework based on the combinations of three baseline EMO algorithms and two batch recommendations mechanisms. Extensive experiments on $136$ benchmark test problem instances with irregular PFs fully demonstrate the effectiveness and superiority of our proposed algorithms against six state-of-the-art data-driven EMO algorithms. In particular, our proposed framework is featured with a faster convergence and a stronger resilience to various PF shapes.

The rest of this paper is organized as follows. \pref{sec:preliminaries} first gives some essential preliminaries including definitions pertinent to this paper along with a pragmatic overview of the existing developments in data-driven EMO. \pref{sec:proposal} delineates the technical details of our proposed framework step by step. The experimental results are presented and analyzed in~\pref{sec:experiments}. At the end, \pref{sec:conclusions} concludes this paper and sheds some lights on potential future directions.

% !tex root = main.tex

\section{Preliminaries}
\label{sec:preliminaries}

In this section, we first give some basic concepts pertinent to this paper. Thereafter, we briefly overview some selected developments of data-driven EMO.

\subsection{Basic Definitions in Multi-Objective Optimization}
\label{sec:definitions}

The MOP considered in this paper is defined as:
\begin{equation}
    \begin{array}{l l}
        \mathrm{minimize} \quad \mathbf{F}(\mathbf{x})=(f_{1}(\mathbf{x}),\cdots,f_{m}(\mathbf{x}))^{T}\\
        \mathrm{subject\ to} \;\; \mathbf{x} \in\Omega%g_j(\mathbf{x})|_{j=1}^k\leq 0\\%,\quad j\in\{1,\cdots,k\}\\
        %\mathrm{\ } \quad\quad\quad\quad\;\; \mathbf{x} \in\Omega
    \end{array},
    \label{eq:MOP}
\end{equation}
where $\mathbf{x}=(x_1,\cdots,x_n)^T$ is a decision vector and $\mathbf{F}(\mathbf{x})$ is an objective vector. $\Omega=[x_i^L,x_i^U]^n\subseteq\mathbb{R}^n$ defines the search space. $\mathbf{F}: \Omega\rightarrow\mathbb{R}^m$ is the corresponding attainable set in the objective space $\mathbb{R}^m$.
\begin{definition}
    Given two solutions $\mathbf{x}^1,\mathbf{x}^2\in\Omega$, $\mathbf{x}^1$ is said to \underline{dominate} $\mathbf{x}^2$, denoted as $\mathbf{x}^1\preceq\mathbf{x}^2$, if and only if $f_i(\mathbf{x}^1)\leq f_i(\mathbf{x}^2)$ for all $i\in\{1,\cdots,m\}$ and $\mathbf{F}(\mathbf{x}^1)\neq\mathbf{F}(\mathbf{x}^2)$. 
\end{definition}
\begin{definition}
    A solution $\mathbf{x}^\ast\in\Omega$ is said to be \underline{Pareto-optimal} if and only if $\nexists\mathbf{x}^\prime\in\Omega$ such that $\mathbf{x}^\prime\preceq\mathbf{x}^\ast$.
\end{definition}
\begin{definition}
    The set of all Pareto-optimal solutions is called the \underline{Pareto-optimal set} (PS), i.e., $PS=\{\mathbf{x}^\ast|\nexists\mathbf{x}^\prime\in\Omega \text{ such that } \mathbf{x}^\prime\preceq\mathbf{x}^\ast\}$ and their corresponding objective vectors form the \underline{Pareto-optimal front} (PF), i.e., $PF=\{\mathbf{F}(\mathbf{x}^\ast)|\mathbf{x}^\ast\in PS\}$.
\end{definition}
\begin{theorem}[\underline{KKT conditions}~\cite{KuhnT51}]
    Let $\mathbf{x}^\ast$ be a Pareto-optimal solution of the MOP with $k$ constraints $\{g_i(\mathbf{x})\leq 0\}_{i=1}^k$ and the set of vectors $\{\nabla g_j(\mathbf{x}^\ast)|j\ \text{is the index of an active constraint}\}$ are linearly independent. There exists vectors $\alpha=(\alpha_1,\cdots,\alpha_m)^T\in\mathbb{R}^m$ and $\lambda=(\lambda_1,\cdots,\lambda_k)^T\in\mathbb{R}^k$, such that:
    \begin{equation}
        \begin{aligned}
            \sum_{i=1}^m\alpha_i\nabla f_i(\mathbf{x}^\ast)+\sum_{j=1}^k\lambda_j\nabla g_j(\mathbf{x}^\ast)&=0\\
            \lambda_jg_j(\mathbf{x}^\ast)|_{j=1}^k&=0
        \end{aligned},
        \label{eq:kkt}
    \end{equation}
    where $\alpha_i\geq 0$, $\forall i\in\{1,\cdots,m\}$ and $\sum_{i=1}^m\alpha_i=1$.
\end{theorem}

\begin{remark}
    The objective and constraint functions are assumed to be continuously differentiable in the KKT conditions.
\end{remark}

\begin{remark}
    The MOP~(\ref{eq:MOP}) considered in this paper does not consider constraints, thus we ignore the $\sum_{j=1}^k\lambda_j\nabla g_j(\mathbf{x}^\ast)$ part of~\pref{eq:kkt} in the latter derivations.
\end{remark}

\begin{corollary}
    The PF is a $(m-1)$-dimensional piecewise continuous manifold under the KKT conditions. For any solution $\mathbf{x}^\ast$ in the PS, there exists an open neighborhood $\Xi(\mathbf{x}^\ast)$ such that the intersection of the PF and $\{\mathbf{F}(\tilde{\mathbf{x}})|\tilde{\mathbf{x}}\in\Xi(\mathbf{x}^\ast)\}$ is a $(m-1)$-dimensional continuously differentiable manifold in $\mathbb{R}^m$~\cite{Hillermeier01}, so as the PS.
    \label{corollary:manifold}
\end{corollary}

\begin{definition}
    Let $\mathcal{M}$ be a continuously differentiable manifold, $\gamma:(-\epsilon,\epsilon)\rightarrow\mathcal{M}$ be a continuously differentiable curve on this manifold and it passes through $\mathbf{x}\in\mathcal{M}$ where $\epsilon>0$. Use $t\in(-\epsilon,\epsilon)$ to parameterize $\gamma$ as $\gamma(t)$ where $\gamma(0)=\mathbf{x}$, the \underline{tangent vector} of $\gamma(0)$, denoted as $\mathbf{v}$, is defined as:
    \begin{equation}
        \left.\mathbf{v}=\frac{\mathrm{d}}{\mathrm{d}t}f\circ\gamma(t)\right|_{t=0},
    \end{equation}
    where $f\circ\gamma(t): (-\epsilon,\epsilon)\rightarrow\mathcal{M}\rightarrow\mathbb{R}$ is a composite mapping.
    \label{definition:tangent_vector}
\end{definition}

\begin{definition}
    The set of all tangent vectors at $\mathbf{x}$ is called the \underline{tangent space} of $\mathcal{M}$ at $\mathbf{x}$, denoted as $\mathcal{T}_{\mathbf{x}}\mathcal{M}$.
\end{definition}

\begin{theorem}
    Let $\mathcal{M}$ be a smooth manifold and $\mathbf{x}\in\mathcal{M}$, then $\dim(\mathcal{T}_{\mathbf{x}}\mathcal{M})=\dim(\mathcal{M})$, where $\dim(\cdot)$ returns the corresponding dimensionality.
\end{theorem}

\begin{figure}[t!]
    \centering
    \includegraphics[width=.6\linewidth]{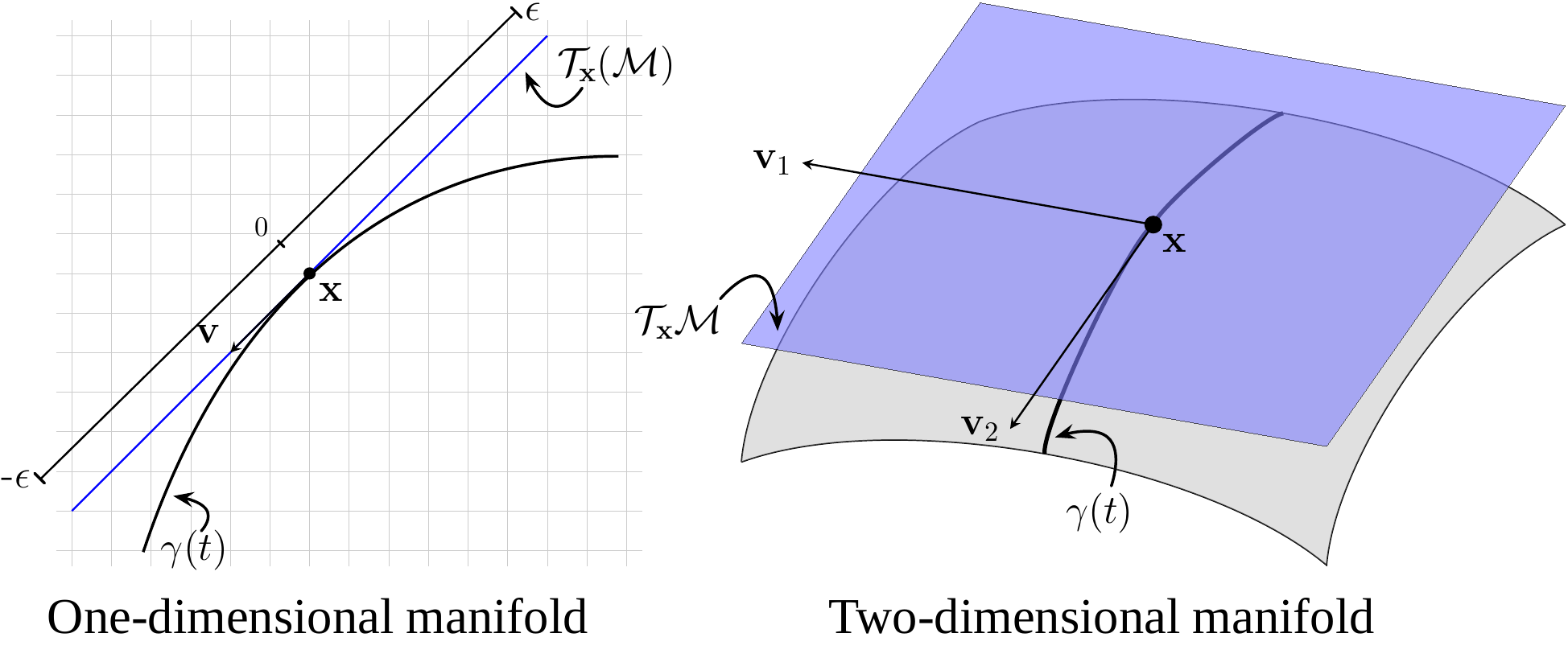}
    \caption{A conceptual illustration of the tangent vector(s) $\mathbf{v}$ of a point $\mathbf{x}$ on a manifold along with its corresponding tangent space $\mathcal{T}_{\mathbf{x}}\mathcal{M}$.}
    \label{fig:tangent_space}
\end{figure}

\begin{remark}
    \pref{fig:tangent_space} gives a conceptual illustration of the tangent vector(s) $\mathbf{v}$ of a point $\mathbf{x}$ on a one- and a two-dimensional manifold, respectively, along with its corresponding tangent space $\mathcal{T}_{\mathbf{x}}\mathcal{M}$. In particular, the number of tangent vectors is $\dim(\mathcal{T}_\mathbf{x}\mathcal{M})$.
\end{remark}

\subsection{Gaussian Process Regression Model}
\label{sec:GP}

Given a set of training data $\mathcal{D}=\{(\mathbf{x}^i,f(\mathbf{x}^i)\}_{i=1}^{N}$, a GPR model aims to learn a latent function $g(\mathbf{x})$ by assuming $f(\mathbf{x}^i)=g(\mathbf{x}^i)+\epsilon$ where $\epsilon\sim\mathcal{N}(0,\sigma^2_n)$ is an independently and identically distributed Gaussian noise. For each testing input vector $\mathbf{z}^\ast\in\Omega$, the mean and variance of the target $f(\mathbf{z}^\ast)$ are predicted as~\cite{RasmussenW06}:
\begin{align}
\overline{g}(\mathbf{z}^\ast)&=m(\mathbf{z}^\ast)+{\mathbf{k}^\ast}^T(K+\sigma_n^2 I)^{-1}(\mathbf{f}-\mathbf{m}(X)),\\
\mathbb{V}[g(\mathbf{z}^\ast)]&=k(\mathbf{z}^\ast,\mathbf{z}^\ast)-{\mathbf{k}^\ast}^T (K+\sigma_n^2 I)^{-1} {\mathbf{k}^\ast},
\label{eq:GP}
\end{align}
where $X=(\mathbf{x}^1,\cdots,\mathbf{x}^N)^T$ and $\mathbf{f}=(f(\mathbf{x}^1),\cdots,f(\mathbf{x}^N))^T$. $\mathbf{m}(X)$ is the mean vector of $X$, $\mathbf{k}^\ast$ is the covariance vector between $X$ and $\mathbf{z}^\ast$, and $K$ is the covariance matrix of $X$. In particular, a covariance function, also known as a kernel, is used to measure the similarity between a pair of data samples $\mathbf{x}$ and $\mathbf{x}^\prime\in\Omega$. This paper uses the Mat\'ern $5/2$ kernel without loss of generality and it is defined as:
\begin{equation}
    k(\mathbf{x},\mathbf{x}')=\sigma_n^2\left(1+\frac{\sqrt{5}d}{\rho}+\frac{5d^2}{3\rho^2}\right)\exp\left(-\frac{\sqrt{5}d}{\rho}\right),
\end{equation} 
where $\rho$ is a positive hyper-parameter of the covariance function and $d=\sqrt{(\mathbf{x}-\mathbf{x}^\prime)^T\cdot(\mathbf{x}-\mathbf{x}^\prime)}$ is the Euclidean distance between $\mathbf{x}$ and $\mathbf{x}^\prime$. The predicted mean $\overline{g}(\mathbf{z}^\ast)$ is directly used as the prediction of $f(\mathbf{z}^\ast)$, and the predicted variance $\mathbb{V}[g(\mathbf{x}^\ast)]$ quantifies the uncertainty. As recommended in~\cite{RasmussenW06}, the hyperparameters are learned by maximizing the log marginal likelihood function defined as: 
\begin{equation}
    \begin{aligned}
        \log p(\mathbf{f}|X)&=-\frac{1}{2}(\mathbf{f}-\mathbf{m}(X))^T(K+\sigma_n^2 I)^{-1}(\mathbf{f}-\mathbf{m}(X))\\
                             &-\frac{1}{2}\log|K+\sigma_n^2 I|-\frac{N}{2}\log 2\pi.
    \end{aligned}
    \label{eq:likelihood}
\end{equation}
In this paper, we assume that the mean function is a constant $0$ and the inputs are noiseless.
%$\nu$ and $\ell$ are positive parameters, $d(\cdot)$ is the Euclidean distance, $K_\nu$ is a modified Bessel function and $\Gamma(\cdot)$ is the gamma function. 

\subsection{Related Works}
\label{sec:related}

This section provides a pragmatic overview of the current developments of data-driven EMO. Interested readers are referred to some survey papers for details~\cite{Jin05,Jin11,JinWCGM19}.

ParEGO~\cite{Knowles06} is one of the earliest attempts that extends the classic efficient global optimization (EGO) algorithm to the context of multi-objective optimization. During each iteration, it randomly generates a weight vector to constitute a scalarizing function of the original MOP. It uses a Kriging model to fit a surrogate model of the underlying scalarizing function, based on which an EA is used to search for the next point of merit by optimizing the expected improvement. In~\cite{EmmerichGN06}, Emmerich et al. proposed to use Hypervolume measure as an alternative of scalarizing function to derive a couple of acquisition functions for multi-objective EGO. The similar idea is further exploited in~\cite{PonweiserWBV08} and~\cite{WagnerEDP10}. Later, Zhang et al.~\cite{ZhangLTV10} proposed a MOEA/D version of EGO, dubbed MOEA/D-EGO. It applies the GPR to fit a surrogate model for each expensive objective function, based on which they derived the estimated mean and variance of the corresponding scalarizing function. Then, a regular MOEA/D routine is used to search for the approximated PF. In addition, they developed a batch recommendation mechanism to pick up more than one candidate solution for expensive FEs at the end of each iteration. In~\cite{ChughJMHS18}, K-RVEA is proposed for expensive many-objective optimization problems. To tackle the problems with irregular PFs, Habib et al.~\cite{HabibSCRM19} proposed HSMEA that takes advantages of the interplay of multiple surrogate models and two sets of reference vectors. In addition, it applies a local search to further exploit high quality infill solutions. To have a well balance between exploration and exploitation, Wang et al. proposed to tune the hyperparameters of the acquisition function in EGO according to the search dynamics on the fly~\cite{WangJSO20}.

In addition to EGO, some other machine learning models have also been studied in data-driven EMO. For example, Voutchkov and Keane~\cite{VoutchkovK10} proposed a simple idea to directly apply a GPR model to replace the expensive objective functions in NSGA-II. At the end of each iteration, the current best candidate solutions in terms of ranking and space filling properties are chosen for conducting expensive FEs. In view of the high computational complexity of GPR, Guo et al.~\cite{GuoJDC19} proposed a heterogeneous ensemble model based on least square SVM and RBFN for surrogate modeling. To identify the infill solution(s) for expensive FEs, an ensemble generation method is proposed to quantify the uncertainty of sample points. In~\cite{SunLGHL11,MartinezC13,AkhtarS16}, RBFN are used as the surrogate model to drive the search process. Instead of a regression model, Pan~\cite{PanHTWZJ19} and Zhang et al.~\cite{ZhangZZ15} proposed to use a classification model to drive the surrogate search routine. Differently, Loshchilov~\cite{LoshchilovSS10} and Seah et al.~\cite{SeahOTJ12} proposed to fit a surrogate model that predicts the Pareto dominance relation between pairs of solutions.

Different from the above mentioned works, another emerging area is to use transfer learning techniques to boost the search process. For example, Luo et al.~\cite{LuoGOW19} proposed to use a multi-task GPR model to build multiple surrogate models simultaneously for different subregions of the PF. In addition, a new infill criterion based on the surrogate landscape is proposed to determine the next candidate solution for conducting the expensive FE. Min et al.~\cite{MinOGG19} proposed to use the transfer stacking technique to jump start the underlying problem-solving routine by leveraging the model built for other related problems. In~\cite{YangDJC20}, Yang et al. proposed an EA assisted by two surrogate models. One model aims to guide the algorithm to quickly find a promising subregion in the search space and the other one focuses on leveraging good solutions according to the knowledge transferred from the first model.

In the classic multi-objective optimization literature, the KKT conditions have been applied to solve bi-objective design optimization problems~\cite{RakowskaHW91}. Later, this idea was generalized to MOPs with any number of objectives in theory~\cite{Hillermeier01,SchulzWGSM18}. It is worth noting that all these approaches are developed upon the assumption that the objective functions are analytically accessible and differentiable. In addition, they only considered a local expansion of an known Pareto-optimal solution. Another line of research is~\cite{DebA16} that developed a proximity measure based on KKT optimality theory. This measure was originally designed to evaluate the convergence of a set of non-dominated solutions with regard to the PS. Later, it has also been used as either a driving force or a termination criterion of a local search procedure in NSGA-III~\cite{DebAS17,AbouhawwashSD17,SeadaAD19}

% !tex root = main.tex

\section{Proposed Algorithm}
\label{sec:proposal} 

\begin{figure*}[t!]
\centering
\includegraphics[width=\linewidth]{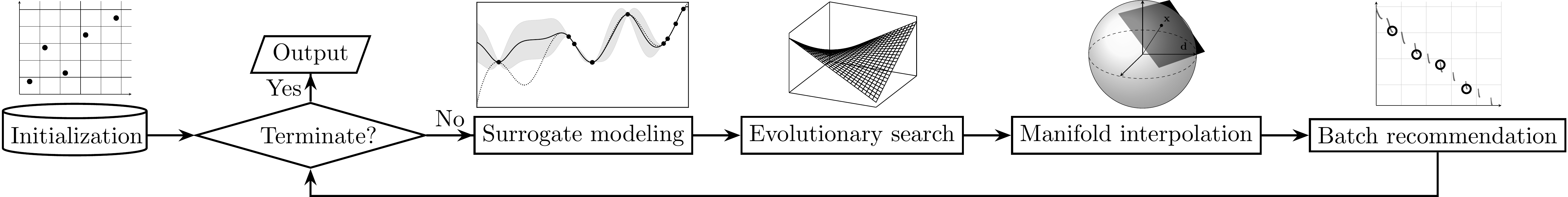}
\caption{Flowchart of the our proposed batched data-driven EMO framework.}
\label{fig:flowchart}
\end{figure*}

The flowchart of our proposed batched data-driven EMO framework based on manifold interpolation is shown in~\pref{fig:flowchart}. It starts with an \texttt{initialization} step based on an experimental design method~\cite{SantnerWN18}. In this paper, we use the classic Latin hypercube sampling to serve this purpose without loss of generality. Then, we evaluate the objective function values of these initial samples and store them in the training dataset. During the main loop, the \texttt{surrogate modeling} step builds a surrogate model for each expensive objective function based on the up-to-date training dataset. In particular, we apply the GPR as the surrogate model in view of its continuously differentiable characteristics. As for the other three steps, we will delineate their implementations in the following paragraphs.

\subsection{Evolutionary Search}
\label{sec:evolutionary_search}

The \texttt{evolutionary search} step aims to approximate the PF based on the surrogate model built in the \texttt{surrogate modeling} step. We argue that any existing EMO algorithm can be used as the surrogate optimizer in this step. In particular, we directly use the surrogate model to replace the expensive objective functions in the EMO algorithm. This paper applies three iconic EMO algorithms, i.e., NSGA-II, IBEA and MOEA/D, for a proof-of-concept purpose. For the sake of being self-contained, we briefly describe their working mechanisms as follows.

\subsubsection{NSGA-II}
\label{sec:nsga2}

This is one of most popular dominance-based EMO algorithms in the literature. It uses the Pareto dominance to promote the convergence and the crowding distance to maintain the population diversity. The general working mechanism of NSGA-II is given as follows.

\begin{enumerate}[Step 1:]
    \item Initialize a population of solutions $\mathcal{P}=\{\mathbf{x}^i\}_{i=1}^N$.
    \item Use crossover and mutation to generate a population of offspring $\mathcal{Q}$.
    \item Use non-dominated sorting~\cite{DebAPM02} to divide $\mathcal{R}=\mathcal{P}\bigcup\mathcal{Q}$ into several non-domination fronts $\mathcal{F}_1,\mathcal{F}_2,\cdots$.
    \item Starting from $\mathcal{F}_1$, solutions are stored in a temporary archive $\overline{\mathcal{P}}$ till its size for the first time equals or exceeds $N$, where $\overline{\mathcal{P}}=\bigcup_{i=1}^{\ell}\mathcal{F}_i$. In particular, $\mathcal{F}_{\ell}$ is the last acceptable non-domination front. If the size of $\overline{\mathcal{P}}$ equals $N$, then let $\mathcal{P}=\overline{\mathcal{P}}$ and go to Step 6; otherwise go to Step 5.
    \item Calculate the crowding distance of solutions in $\mathcal{F}_\ell$ and sort them in a descending order. Remove the last $|\overline{\mathcal{P}}|-|\mathcal{P}|$ solutions from $\overline{\mathcal{P}}$ and let $\mathcal{P}=\overline{\mathcal{P}}$.
    \item If the stopping criterion is met, then stop and output $\mathcal{P}$. Otherwise, go to Step 2.
\end{enumerate}

\subsubsection{IBEA}
\label{sec:ibea}

The basic idea of IBEA is to transform a MOP into a single-objective optimization problem in terms of a binary performance indicator. Then it directly uses this indicator in the environmental selection process. The general working mechanism of IBEA is given as follows.
\begin{enumerate}[Step 1:]
    \item Initialize a population of solutions $\mathcal{P}=\{\mathbf{x}^i\}_{i=1}^N$.
    \item Use crossover and mutation to generate a population of offspring $\mathcal{Q}$ and let $\overline{\mathcal{P}}=\mathcal{P}\bigcup\mathcal{Q}$.
    \item While $|\overline{\mathcal{P}}|>N$ do
    \begin{enumerate}[Step 3.1:]
        \item Find the solution $\mathbf{x}^\ast=\argmin_{\mathbf{x}\in\overline{\mathcal{P}}}F(\mathbf{x})$ and remove it from $\overline{\mathcal{P}}$, i.e., $\overline{\mathcal{P}}=\overline{\mathcal{P}}\setminus\{\mathbf{x}^\ast\}$.
        \item Update the fitness values of solutions in $\overline{\mathcal{P}}$, i.e., $\forall\mathbf{x}\in\overline{\mathcal{P}}$, $F(\mathbf{x})=F(\mathbf{x})+e^{-\mathtt{I}(\{\mathbf{x}^\ast\},\{\mathbf{x}\})/\kappa}$.
    \end{enumerate}
    Let $\mathcal{P}=\overline{\mathcal{P}}$ and go to Step 4.
    \item If the stopping criterion is met, then stop and output $\mathcal{P}$. Otherwise, go to Step 2.
\end{enumerate}

\begin{remark}
The fitness value of a solution $\mathbf{x}$ is calculated as:
\begin{equation}
F(\mathbf{x})=\sum_{\mathbf{x}^\prime\in\overline{\mathcal{P}}\setminus\{\mathbf{x}\}}-e^{-\mathtt{I}(\{\mathbf{x}^\prime\},\{\mathbf{x}\})/\kappa},
\end{equation}
where $\kappa$ is a user defined scaling factor and we set $\kappa=0.05$~\cite{ZitzlerK04}. $\mathtt{I}(\cdot,\cdot)$ is a binary quality indicator and we use the $\mathtt{I}_{\mathtt{HD}}$-indicator as in~\cite{ZitzlerK04} based on the Hypervolume indicator:
\begin{equation}
\mathtt{I}_{\mathtt{HD}}(\mathcal{A},\mathcal{B})=
\begin{dcases}
\mathtt{I}_{\mathtt{H}}(\mathcal{B})-\mathtt{I}_{\mathtt{H}}(\mathcal{A}) & \text{if}\ \forall\mathbf{x}^\prime\in\mathcal{B}, \exists\mathbf{x}\in\mathcal{A}: \mathbf{x}\preceq\mathbf{x}^\prime \\
\mathtt{I}_{\mathtt{H}}(\mathcal{A}+\mathcal{B})-\mathtt{I}_{\mathtt{H}}(\mathcal{A}) & \text{otherwise}
\end{dcases},
\end{equation}
where $\mathtt{I}_{\mathtt{H}}(\mathcal{A})$ is the Hypervolume of the objective space dominated by $\mathcal{A}$ and $\mathtt{I}_{\mathtt{HD}}(\mathcal{A},\mathcal{B})$ evaluates the Hypervolume of the space that is dominated by $\mathcal{B}$ but not by $\mathcal{A}$.
\end{remark}

\subsubsection{MOEA/D}
\label{sec:moead}

The basic idea of MOEA/D is to decompose the original MOP into several subproblems and it uses a population-based technique to solve these subproblems in a collaborative manner. Given a weight vector $\mathbf{w}$, this paper uses the Tchebycheff function~\cite{LiZKLW14} as a subproblem:
\begin{equation}
    \begin{array}{l l}
        \mathrm{minimize~}\quad g(\mathbf{x}|\mathbf{w},\mathbf{z}^\ast)&=\max_{1\leq i\leq m}{\frac{|f_i(\mathbf{x})-z_i^\ast|}{w_i}}\\
        \mathrm{subject\ to}\quad \mathbf{x}\in\Omega
    \end{array},
    \label{eq:TCH}
\end{equation} 
where $\mathbf{z}^\ast$ is the ideal point. The general working mechanism of MOEA/D is given as the following three-step process.
\begin{enumerate}[Step 1:]
    \item Initialize a population of solutions $\mathcal{P}=\{\mathbf{x}^i\}_{i=1}^N$, a set of weight vectors $\mathcal{W}=\{\mathbf{w}^i\}_{i=1}^N$ and their neighborhood structure. Randomly assign each solution to a weight vector.
    \item For $i=1,\cdots,N$, do%each reference point, mating parents are selected from its neighborhood.
        \begin{enumerate}[Step 2.1:]
            \item Randomly select a required number of mating parents from $\mathbf{w}^i$'s neighborhood.
            \item Use crossover and mutation to reproduce an offspring $\mathbf{x}^c$.
            \item Use $\mathbf{x}^c$ to update the subproblems within the neighborhood of $\mathbf{w}^i$.
        \end{enumerate}
    \item If the stopping criterion is met, then stop and output the population. Otherwise, go to Step 2.
\end{enumerate}
\begin{remark}
In Step 1, we use the Das and Dennis's method~\cite{NBI} to initialize a set of evenly distributed weight vectors from a canonical simplex. The neighborhood structure $B(i)$ of each weight vector $\mathbf{w}^i$, $i\in\{1,\cdots,N\}$, contains its $T$ closest weight vectors, where $T=20$ as suggested in~\cite{LiZ09}. %\pref{fig:weights} gives two examples of reference point distribution and the neighbourhood of a reference point in the two- and three-objective cases.
%\begin{figure}[htbp]
%\centering
%\subfloat[2-D case.]{\includegraphics[width=.5\linewidth]{figs/weight2D.pdf}}
%\subfloat[3-D case.]{\includegraphics[width=.5\linewidth]{figs/weight3D.pdf}}
%\caption{Illustration of reference points generated by the Das and Dennis' method~\cite{NBI}. The black circle represents the neighbourhood of a particular reference point.}
%\label{fig:weights}
%\end{figure}
\end{remark}

\begin{remark}
In Step 2.1, to improve the exploration ability, there is a small probability $\delta=0.1$ to select mating parents from the whole population as suggested in~\cite{LiZ09}.
\end{remark}

\begin{remark}
Each subproblem is associated with a unique solution. In Step 2.3, $\mathbf{x}^c$ can update a particular subproblem $\mathbf{w}$ if and only if $g(\mathbf{x}^c|\mathbf{w},\mathbf{z}^\ast)<g(\mathbf{x}|\mathbf{w},\mathbf{z}^\ast)$, where $\mathbf{x}$ is the solution originally associated with $\mathbf{w}$.
\end{remark}

\begin{remark}
In Step 2.3, $\mathbf{x}^c$ has a small probability $\delta=0.1$ to update a subproblem from $\mathcal{W}$, rather than merely in $B(i)$.
\end{remark}

\subsection{Manifold Interpolation}
\label{sec:manifold_interpolation}

After the \texttt{evolutionary search} step, we obtain a population of solutions $\mathcal{P}$ that approximate the surrogate PF. Here we assume that these solutions are Pareto-optimal thus they all satisfy the KKT conditions. According to~\pref{corollary:manifold}, $\forall\mathbf{x}\in\mathcal{P}$, the PS segment with regard to an open neighborhood $\Xi(\mathbf{x})$ is a $(m-1)$-dimensional manifold, denoted as $\mathcal{M}_\mathbf{x}$, so does the corresponding PF segment. The basic idea of this \texttt{manifold interpolation} step is to interpolate a set of $\tilde{N}\gg 1$ new candidate solutions $\mathcal{S}=\{\hat{\mathbf{x}}^i\}_{i=1}^{\tilde{N}}$ along the tangent space of $\mathbf{x}$. More specifically,
\begin{equation}
\hat{\mathbf{x}}=\mathbf{x}+\sum_{i=1}^{m-1}\eta_i\mathbf{v}_i,
\label{eq:interpolation}
\end{equation}
where $\mathbf{v}_i$ is the $i$-th tangent vector at $\mathbf{x}$ and $\eta_i\in(0,1]$ is a random scaling factor along that direction. In the following paragraphs, we will derive a closed form method to calculate the tangent vectors. To facilitate our derivation, as in~\pref{definition:tangent_vector}, we use a parametric form $\mathbf{x}(t)$ where $t\in(-\epsilon,\epsilon)$ to represent each solution on a smooth curve passing through $\mathbf{x}$ on $\mathcal{M}_\mathbf{x}$ where $\mathbf{x}(0)=\mathbf{x}$. 

According to~\pref{corollary:manifold}, we have $\forall\mathbf{x}(t)\in\Xi(\mathbf{x})$ satisfies the KKT conditions. We assume that there exist a time-varying parameter vector $\mathbf{\alpha}(t)=(\alpha_1(t),\cdots,\alpha_m(t))^T\in\mathbb{R}^m$, $t\in(-\epsilon,\epsilon)$, such that:
\begin{equation}
\sum_{i=1}^m\alpha_i(t)\nabla f_i(\mathbf{x}(t))=0,
    \label{eq:kkt_t}
    \end{equation}
    where $\alpha_i(t)\geq 0$ and $\sum_{i=1}^m\alpha_i(t)=1$. $f_i(\mathbf{x}(t))$ is actually a composite mapping $f_i\circ\mathbf{x}(t):(-\epsilon,\epsilon)\rightarrow\mathcal{M}_\mathbf{x}\rightarrow\mathbb{R}$ on the manifold as in~\pref{definition:tangent_vector} where $i\in\{1,\cdots,m\}$. By taking the derivatives of~\pref{eq:kkt_t} at $t=0$, we have:
    \begin{align}
    \left.\frac{d}{dt}\sum_{i=1}^m \alpha_i(t)\nabla f_i(\mathbf{x}(t))\right|_{t=0}&=0,\nonumber \\
        \implies\sum_{i=1}^m\alpha_i^\prime(0)\nabla f_i(\mathbf{x}(0))+(\sum_{i=1}^m\alpha_i(0)\nabla^2f_i(\mathbf{x}(0))\mathbf{x}^\prime(0)&=0.%\nonumber \\
                %\implies(\sum_{i=1}^m\alpha_i(0)\nabla^2f_i(\mathbf{x}(0))\mathbf{x}^\prime(0)=-\sum_{i=1}^m\alpha_i^\prime(0)\nabla f_i(\mathbf{x}(0)).
                    \label{eq:kkt_derivative}
                    \end{align}
                    Given that $\sum_{i=1}^m\alpha_i^\prime(t)=0$, we rewrite~\pref{eq:kkt_derivative} as a system of linear equations:
                    \begin{equation}
                    \begin{bmatrix}\mathbf{1}_{1\times m} & \mathbf{0}_{1\times n}\\ \mathbf{J}^T_{\mathbf{F}(\mathbf{x}(0))} & \mathbf{H}^T_{\mathbf{F}(\mathbf{x}(0))}\cdot\mathbf{\alpha(0)}\end{bmatrix}\begin{bmatrix}\alpha^\prime(0)\\ \mathbf{x}^\prime(0)\end{bmatrix}=0,
                    \label{eq:system}
                    \end{equation}
                    where $\mathbf{J}_{\mathbf{F}(\mathbf{x}(0))}$ and $\mathbf{H}_{\mathbf{F}(\mathbf{x}(0))}$ are the $m\times n$ Jacobian matrix and $m\times m\times n$ Hessian tensor of $\mathbf{F}(\mathbf{x}(0))=(f_1(\mathbf{x}(0)),\cdots,f_m(\mathbf{x}(0)))^T$, respectively. By solving this system of linear equations~(\ref{eq:system}), we obtain $m-1$ different $\mathbf{x}^\prime(0)$, which constitute the $m-1$ tangent vectors $\{\mathbf{v}_i\}_{i=1}^{m-1}$ in~\pref{eq:interpolation}.
\begin{remark}
Let us rewrite~\pref{eq:kkt_derivative} as follows:
\begin{equation}
(\underbrace{\sum_{i=1}^m\alpha_i(0)\nabla^2f_i(\mathbf{x}(0)}_{\mathbf{H}_{\mathbf{F}(\mathbf{x}(0))}\cdot\mathbf{\alpha(0)}})\underbrace{\mathbf{x}^\prime(0)}_{\mathbf{v}}=-\sum_{i=1}^m\alpha_i^\prime(0)\nabla f_i(\mathbf{x}(0)).
\label{eq:kkt_rewrite}
\end{equation}
    The left hand side of~\pref{eq:kkt_rewrite} is thus a linear combination of $\{\nabla f_i(\mathbf{x}(0))\}_{i=1}^m$ that constitute a subspace spanned by them. We take the inverse of $\mathbf{H}_{\mathbf{F}(\mathbf{x}(0))}\cdot\alpha(0)$ and further derive~\pref{eq:kkt_rewrite} as:
    \begin{equation}
        \mathbf{x}^\prime(0)=\Big[\mathbf{H}_{\mathbf{F}(\mathbf{x}(0))}\cdot\alpha(0)\Big]^{-1}\Big[-\sum_{i=1}^m\alpha_i^\prime(0)\nabla f_i(\mathbf{x}(0))\Big].
    \end{equation}
%    \pref{fig:derivative_example} gives an illustative example of a two-objective problem. In particular, since the tangent vector $\mathbf{v}$ navigate the solution $\mathbf{x}(0)$ through the PS manifold, it accordingly interpolates new Pareto-optimal solutions along the PF.
\end{remark}

\begin{figure}[t!]
    \centering
    \includegraphics[width=.6\linewidth]{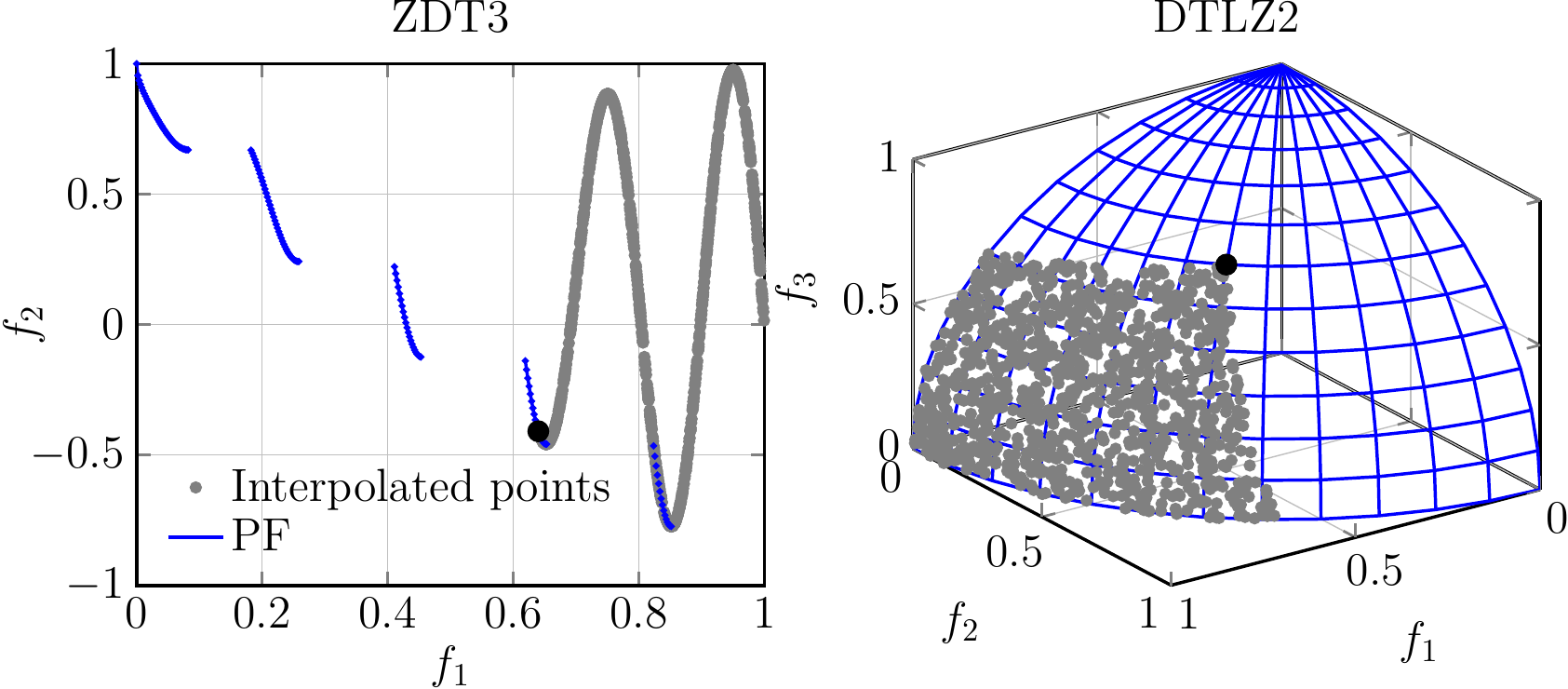}
    \caption{Examples of interpolated solutions (denoted as grey circles) generated by the manifold interpolation.}
    \label{fig:manifold_interpolation}
\end{figure}

\begin{remark}
    As shown in~\pref{eq:interpolation}, this \texttt{manifold interpolation} step implements a random walk along the tangent space of $\mathbf{x}$. In principle, the generated solutions constitute a piece-wise linear approximation to the corresponding PS and PF segments within a neighborhood. \pref{fig:manifold_interpolation} gives two examples of manifold interpolation at a given point on the 2-objective ZDT3 and the 3-objective DTLZ2.
\end{remark}

To constitute the Jacobian matrix and the Hessian tensor used in~\pref{eq:system}, we need to access the first- and second-order derivatives of the underlying objective functions. In this paper, since the objective functions are modeled by the GPR, which is continuously differentiable, we can naturally derive the first- and second-order derivatives of the predicted mean function with regard to a candidate solution $\mathbf{x}$ as:
\begin{equation}
    \frac{\partial \overline{g}(\mathbf{x})}{\partial \mathbf{x}}=\frac{\partial \mathbf{k}^\ast}{\partial \mathbf{x}}K^{-1}\mathbf{f},\quad
    \frac{\partial^2 \overline{g}(\mathbf{x})}{\partial \mathbf{x}^{2}}=\frac{\partial^2 \mathbf{k}^\ast}{\partial \mathbf{x}^{2}}K^{-1}\mathbf{f},
%    \frac{\partial \mathbb{V}[g(\mathbf{z}^\ast)]}{\partial \mathbf{z}^\ast} &= -\frac{\partial \mathbf{k}^\ast}{\partial \mathbf{z}^\ast}K^{-1}\mathbf{k}^{\ast T}-\mathbf{k}^\ast K^{-1}\frac{\partial \mathbf{k}^{\ast T}}{\partial \mathbf{z}^\ast}, 
\end{equation}
where the first- and second-order derivatives of $\mathbf{k}^\ast$, i.e., the covariance vector between $\mathcal{P}$ and $\mathbf{x}$, are calculated as:
\begin{align}
    \frac{\partial \mathbf{k}^\ast}{\partial \mathbf{x}} &= -\frac{5\mathbf{d}}{3\rho^2}\left(1+\frac{\sqrt{5}\mathbf{d}}{\rho}\right)\sigma_n^2\exp\left(-\frac{\sqrt{5}\mathbf{d}}{\rho}\right)\frac{\partial \mathbf{d}}{\partial \mathbf{x}},\\
    \frac{\partial^2 \mathbf{k}^\ast}{\partial \mathbf{x}^2} &= -\frac{5}{3\rho^2}\left(1+\sqrt{5}\mathbf{d}-\frac{5\mathbf{d}^2}{\rho^2}\right)\sigma_n^2\exp\left(-\frac{\sqrt{5}\mathbf{d}}{\rho}\right)\left(\frac{\partial \mathbf{d}}{\partial \mathbf{x}}\right)^2\\\nonumber & -\frac{5\mathbf{d}}{3\rho^2}\left(1+\frac{\sqrt{5}\mathbf{d}}{\rho}\right)\sigma_n^2\exp\left(-\frac{\sqrt{5}\mathbf{d}}{\rho}\right)\frac{\partial^2 \mathbf{d}}{\partial \mathbf{x}^2},
\end{align}
where $\mathbf{d}$ is the vector of distances between $\mathcal{P}$ and $\mathbf{x}$. In summary, the working mechanism of this \texttt{manifold interpolation} step is given as follows.
\begin{enumerate}[Step 1:]
\item Initialize the candidate solution set $\mathcal{S}=\emptyset$.
\item For $i=1,\cdots,N$, do
\begin{enumerate}[Step 2.1:]
\item Calculate the tangent vectors of $\mathbf{x}^i\in\mathcal{P}$ by solving the system of linear equations given in~\pref{eq:system}.
\item Use~\pref{eq:interpolation} to generate a set of $\overline{N}=\frac{\tilde{N}}{N}$ candidate solutions $\overline{\mathcal{S}}=\{\hat{\mathbf{x}}^k|\hat{\mathbf{x}}^k=\mathbf{x}_i+\sum_{j=1}^{m-1}\eta_j\mathbf{v}_j\ \text{and}\ \eta_j\in(0,1]\}_{k=1}^{\overline{N}}$.
\item Remove invalid solutions in $\overline{\mathcal{S}}$ outside of $\Omega$.
\item $\mathcal{S}=\mathcal{S}\bigcup\overline{\mathcal{S}}$.
\end{enumerate}
\item Use the GPR model to predict the objective function values of solutions in $\mathcal{S}$.
\item Output the non-dominated solutions in $\mathcal{S}$.
%\item If $|C|<\hat{N}$, go to Step 2; otherwise output $C$.
\end{enumerate}

\subsection{Batch Recommendation}
\label{sec:batch_recommendation}

This step is also known as the infill criterion in the surrogate-assisted EA literature. It aims to pick up $\xi\geq 1$ promising solutions from $\mathcal{C}=\mathcal{P}\bigcup\mathcal{S}$ and evaluate them by using the expensive objective functions. These newly evaluated solutions are then used to update the training dataset for the next iteration. Different from most, if not all, works using GPR as the surrogate model, our infill criterion does not rely on an uncertainty quantification measure, also known as acquisition function in the Bayesian optimization literature~\cite{abs-1807-02811}. Furthermore, selecting a batch of samples to evaluate can significantly reduce the overhead for surrogate modeling. More specifically, we propose two alternative ways to implement this \texttt{batch evaluation} step.
\begin{itemize}
    \item The first one is based on the individual Hypervolume contribution (IHV), independent of the underlying baseline algorithm. We calculate the IHV of each candidate solution $\mathbf{x}\in\mathcal{C}$ as:
    \begin{equation}
        \mathtt{IHV}(\mathbf{x})=\mathtt{HV}(\mathcal{C})-\mathtt{HV}(\mathcal{C}\setminus\{\mathbf{x}\}),
    \end{equation}
where $\mathtt{HV}(\mathcal{C})$ evaluates the Hypervolume (HV)~\cite{ZitzlerT99} of $\mathcal{C}$. Then, the top $\xi$ solutions in $\mathcal{C}$ with the largest IHV are picked up for the expensive evaluations.

    \item The other one is directly derived from the native environmental selection of the baseline EMO algorithm used in the \texttt{evolutionary search} step.
    \begin{itemize}
        \item If the baseline algorithm is NSGA-II, we propose a four-step process for the batch recommendation.
        \begin{enumerate}[Step 1:]
            \item Identify the non-dominated solutions in $\mathcal{C}$ and store them in $\overline{\mathcal{C}}$.
            \item Use $N$ evenly distributed weight vectors to divide the objective space into $N$ subregions and associate each solution in $\overline{\mathcal{C}}$ to its closest weight vector with the smallest acute angle.
            \item Pick up the best solution with the largest crowding distance for each subregion to constitute $\tilde{\mathcal{C}}$.
            \item Pick up the top $\xi$ solutions from $\tilde{\mathcal{C}}$ with the largest crowding distances.
        \end{enumerate} 
%            first use the non-dominated sorting to identify the current non-dominated solutions. Then, the top $\xi$ solutions in $\mathcal{C}$ having the largest crowding distances are picked up for the expensive evaluations.
        \item If the baseline algorithm is IBEA, we can directly use its fitness function to choose $\xi$ best solutions.
        \item If the baseline algorithm is MOEA/D, we propose the following three-step process for the batch recommendation.
        \begin{enumerate}[Step 1:]
            \item For each subproblem $g(\cdot|\mathbf{w}^i,\mathbf{z}^\ast)$ where $i\in\{1,\cdots,N\}$, identify the best solution $\mathbf{x}^{i\ast}$ in $\mathcal{C}$:
            \begin{equation}
                \mathbf{x}^{i\ast}=\argmin_{\mathbf{x}\in\mathcal{C}}g(\mathbf{x}|\mathbf{w}^i,\mathbf{z}^\ast),
            \end{equation}

            \item Calculate the fitness improvement on each subproblem with respect to the previous iteration.
            \begin{equation}
                \Delta_i=\frac{g(\hat{\mathbf{x}}^{i\ast}|\mathbf{w}^i,\mathbf{z}^\ast)-g(\mathbf{x}^{i\ast}|\mathbf{w}^i,\mathbf{z}^\ast)}{g(\hat{\mathbf{x}}^{i\ast}|\mathbf{w}^i,\mathbf{z}^\ast)},
            \end{equation}
where $i\in\{1,\cdots,N\}$ and $\hat{\mathbf{x}}^{i\ast}$ is the best solution of the $i$-th subproblem in the previous iteration.
            \item Pick up the top $\xi$ solutions of which the associated subproblems having the largest fitness improvements.
        \end{enumerate}
    \end{itemize}
\end{itemize}

\subsection{Algorithm Instances}
\label{sec:instances}

As introduced in~\pref{sec:evolutionary_search}, any existing EMO algorithm can be used in the \texttt{evolutionary search} step. With regard to the two batch recommendation methods introduced in~\pref{sec:batch_recommendation}, we propose six algorithm instances for a proof-of-concept purpose, dubbed as \texttt{DMI-x-IHV} by using the IHV for the batch recommendation or \texttt{DMI-x} by using the native environmental selection, respectively, where \texttt{x} is either NSGA-II, IBEA or MOEA/D.

\begin{figure*}[htbp]
\centering
\includegraphics[width=.8\linewidth]{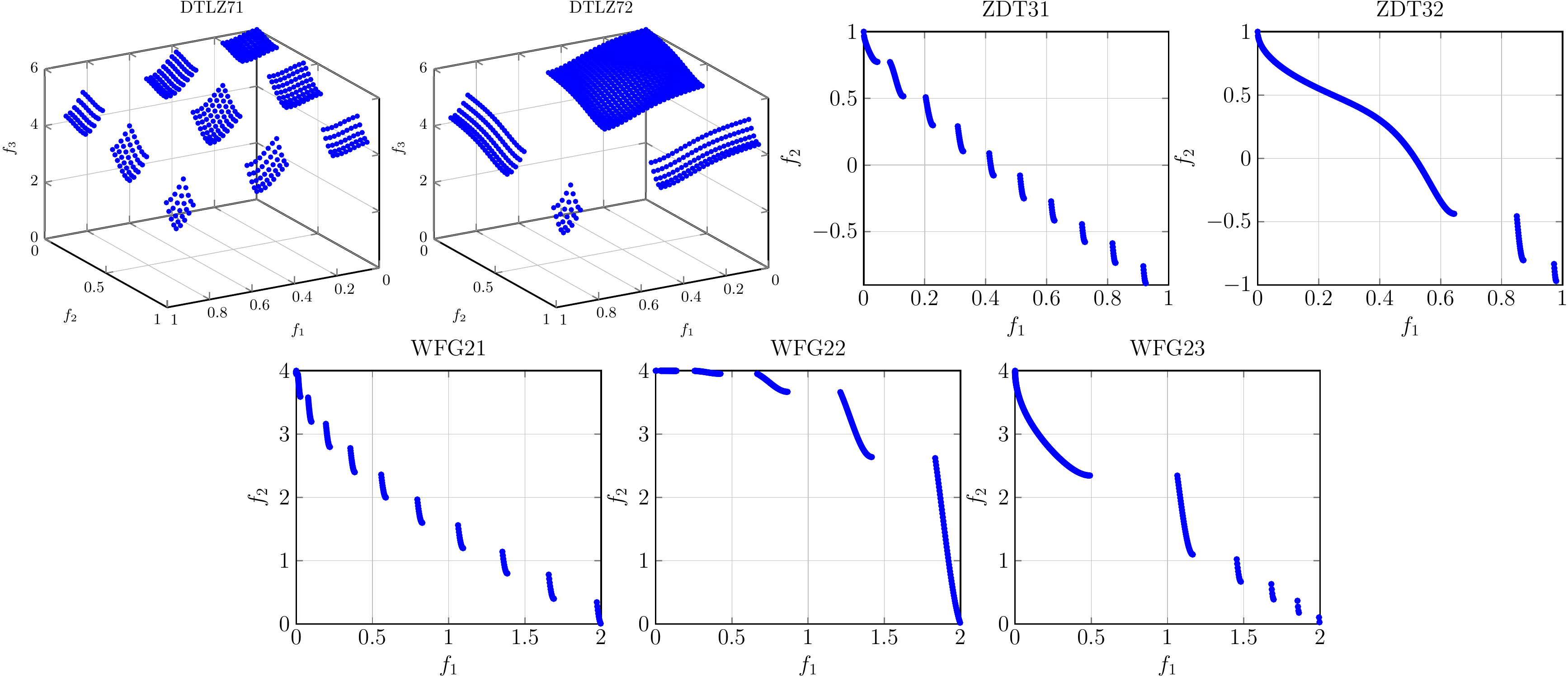}
\caption{Examples of the PFs of our proposed benchmark test problems.}
\label{fig:PFs}
\end{figure*}

\section{Experimental Setup}
\label{sec:settings}

This section introduces the experimental settings for validating the effectiveness of our proposed batched data-driven EMO framework compared against some state-of-the-art algorithms.

\subsection{Benchmark Test Problems}
\label{sec:benchmarks}

In our empirical study, we only consider benchmark test problems with irregular PF shapes to constitute our benchmark suite~\cite{LiZZL09,LiZLZL09,CaoWKL11,LiFK11,CaoKWL12,LiKWCR12,LiKCLZS12,LiKWTM13,LiK14,CaoKWL14,WuKZLWL15,LiKD15,LiDZ15,CaoKWLLK15,LiDZZ17,WuKJLZ17,LiDAY17,LiDY18,LiCSY19,Li19,LiuLC19,GaoNL19,LiXT19,ZouJYZZL19,KumarBCLB18,BillingsleyLMMG19,ChenLY18,LiLDMY20,WuLKZ20,LiX0WT20,LiLLM21,LaiL021,ShanL21,WangYLK21,LiXCT20,LiuLC20}. More specifically, it consists of ZDT3~\cite{ZitzlerDT00}, DTLZ7~\cite{DebTLZ05}, minus DTLZ2~\cite{IshibuchiSMN17}, mDTLZ2~\cite{WangOI19}, WFG2~\cite{HubandHBW06}, WFG41 to WFG48~\cite{WangPF15}. Furthermore, based on ZDT3, DTLZ7 and WFG2, we develop a series of problems with a controllable number disconnected regions and imbalanced sizes. Their mathematical definitions are as follows.
\begin{itemize}
\item ZDT3$\star$
\begin{equation}
\begin{array}{l l}
\mathrm{minimize}\quad  f_1(\mathbf{x})&=x_1 \\ 
\mathrm{minimize}\quad  f_2(\mathbf{x})&=g(\mathbf{x})(1-\sqrt{x_1/g(\mathbf{x})}\\ &\qquad -x_1^\alpha/g(\mathbf{x})\sin{A\pi x_1^\beta}
\end{array},
\end{equation}
where
\begin{equation}
g(\mathbf{x})=1+\frac{9}{n-1}\sum_{i=2}^nx_i.
\end{equation}

\item DTLZ7$\star$
\begin{equation}
    \begin{array}{l l}
        \mathrm{minimize}\quad & f_1(\mathbf{x})=x_1 \\
        & \quad \vdots \\
        \mathrm{minimize}\quad & f_{m-1}(\mathbf{x})=x_{M-1} \\
        \mathrm{minimize}\quad & f_{m}(\mathbf{x})=(1+g(\mathbf{x}_M))h(f_1,\cdots,f_{m-1},g)
        %\text{where}\;& g(\mathbf{x}) = 1 + \frac{9}{|\mathbf{x}_M|} \sum_{x_i\in\mathbf{x}_M}x_i \\
        %& h(f_1,f_2,\dots,f_{M-1},g) = M - \sum_{i=1}^M[\frac{f_i}{1+g}(1+f_i^\alpha \sin(A \pi f_i^\beta))] \\
        %\text{subject to}\;& 0 \leq  x_i \leq 1, \text{for} i = 1,2,\dots,n
    \end{array},
\end{equation}
where 
\begin{equation}
    \begin{array}{l l}
        g(\mathbf{x})&=1+\frac{9}{|\mathbf{x}_m|} \sum_{x_i\in\mathbf{x}_m}x_i \\
        h(f_1,\cdots,f_{m-1},g)&=m\\ & -\sum_{i=1}^m[\frac{f_i}{1+g}(1+f_i^\alpha \sin(A \pi f_i^\beta))]
%        \mathrm{subject\ to}\;& 0 \leq  x_i \leq 1, \text{for} i = 1,2,\dots,n
    \end{array},
\end{equation}
where $x_i\in[0,1]$, $i\in\{1,\cdots,n\}$.

\item WFG2$\star$
\begin{equation}
\begin{array}{l l}
\mathrm{t}^1=\left\{
\begin{array}{ll}
    t^1_{i=1:k} &= x_i \\
    t^1_{k+1:n} &= \mathtt{s\_linear}(x_i,0.35)
\end{array}
\right.\\
    \mathrm{t}^2=\left\{
\begin{array}{ll}
    t^2_{i=1:k} &= t^1_i \\
    t^2_{k+1:n} &= \mathtt{r\_nonsep}(\{t^1_{k+2(i-k)-1},\\ & \qquad t^1_{k+2(i-k)}\},2)
\end{array}
\right.\\
    \mathrm{t}^3=\left\{
\begin{array}{ll}
    t^3_{i=1:m-1} &= \mathtt{r\_sum}(\{t^2_{(i-1)k/(m-1)+1},\\ & \qquad \cdots,t^2_{ik/(m-1)}\},\{1,\cdots,1\})\\
    t^3_{m} &= \mathtt{r\_sum}(\{t^2_{k+1},\cdots,t^2_{k+l/2}\},\\ & \qquad \{1,\cdots,1\})
    \end{array}
\right.\\
\mathrm{shape}=\left\{
\begin{array}{ll}
    f_{1:m-1} &= \mathtt{convex}_m \\ f_m &= \mathtt{disc}_m(A,\alpha,\beta)
\end{array}
\right.
\end{array},
\end{equation}
where the definitions of $\mathtt{s\_linear}(\cdot)$, $\mathtt{r\_nonsep}(\cdot)$, $\mathtt{r\_sum}(\cdot)$, $\mathtt{convex}_m$ and $\mathtt{disc}_m(\cdot)$ can be found in~\cite{HubandHBW06}.

\end{itemize}

Note that the $A$ determines the number of disconnected regions of the PF. $\alpha$ controls the overall shape of the PF where $\alpha>1$, $\alpha<1$ and $\alpha=1$ leads to a concave, a convex and a linear PF, respectively. $\beta$ influences the location of the disconnected regions. In our experiments, we instantiate $7$ test problem instances, the settings used in our experiments are given in~\pref{tab:parameters}. \pref{fig:PFs} gives the illustrative examples of their PFs. The number of objectives is set to $m=2$ for the ZDT and $m=3$ for the DTLZ problems. As for the WFG problems, we consider both 2- and 3-objective cases. The number of variables is set as $n\in\{5,10,20,30\}$ respectively for each benchmark test problem. In total, there are $136$ test problem instances considered in our experiments.

\begin{table}[htbp]
\centering
\caption{Parameter settings of ZDT3$\ast$, DTLZ7$\ast$ and WFG2$\ast$.}
\label{tab:parameters}
\scriptsize
\begin{tabular}{c|c|c|c|c|c|c|c}
\cline{2-8}
                               & ZDT31 & ZDT32 & DTLZ71 & DTLZ72 & WFG21 & WFG22 & WFG23 \\ \hline
\multicolumn{1}{c|}{$A$}      & 10    & 5     & 5      & 3      & 10    & 5     & 5     \\ \hline
\multicolumn{1}{c|}{$\alpha$} & 10    & 0     & 0      & 0      & 1     & 5     & 1     \\ \hline
\multicolumn{1}{c|}{$\beta$}  & 1     & 5     & 1      & 2      & 1     & 1     & 5     \\ \hline
\end{tabular}
\end{table}
 
%\textcolor{red}{In this paper, we consider test problems chosen from five widely used benchmark problem suites including ZDT1 to ZDT4 and ZDT6~\cite{ZitzlerDT00}, DTLZ1 to DTLZ7~\cite{DebTLZ05}, minus DTLZ1 to minus DTLZ4~\cite{IshibuchiSMN17}, mDTLZ1 to mDTLZ4~\cite{WangOI19}, WFG1 to WFG9~\cite{HubandHBW06} and WFG41 to WFG49~\cite{WangPF15}. All these test problems are with continuous variables and have various PF shapes (e.g., linear, convex, concave, disconnected, degenerate and inverted PFs) and different search space properties. They are all scalable to any number of objectives where we set $m=\{3,5,8\}$ except ZDT and mDTLZ problems are constantly with two and three objectives respectively. The number of variables is set as $n=10$ for all test problems.}

\subsection{Peer Algorithms and Parameter Settings}
\label{sec:peers}

To validate the competitiveness of our proposed algorithms, we compare their performance with six state-of-the-art algorithms in the literature, including \texttt{ParEGO}~\cite{Knowles06}, \texttt{MOEA/D-EGO}~\cite{ZhangLTV10}, \texttt{K-RVEA}~\cite{ChughJMHS18}, \texttt{EIM}~\cite{ZhanCL17}, \texttt{TSMEA}~\cite{BradfordSL18} and \texttt{HSMEA}~\cite{HabibSCRM19}. We do not intend to delineate their working mechanisms here while interested readers are referred to their original papers for details.

The parameter settings are listed as follows.
\begin{itemize}
    \item\underline{Number of function evaluations (FEs)}: The initial sampling size is set to $11\times n-1$ for all algorithms and the maximum number of FEs is set as $150$ and $250$ for $m=2$ and $3$, respectively. %As for the algorithm-specific parameters
    \item\underline{Reproduction operators}: The parameters associated with the simulated binary crossover and polynomial mutation are set as $p_c=1.0$, $\eta_c=20$, $p_m=\frac{1}{n}$, $\eta_m=20$. As for those use differential evolution for offspring reproduction, we set $CR=F=0.5$.
    \item\underline{Kriging models}: As for the algorithms that use Kriging for surrogate modeling, the corresponding hyperparameters of the MATLAB Toolbox DACE~\cite{IMM2002-01460} is set to be within the range $[10^{-5},10^5]$.
    \item\underline{Batch size $\xi$}: It is set as $\xi=10$ for our proposed algorithms and $\xi=5$ is set in \texttt{MOEA/D-EGO}, \texttt{K-RVEA} and \texttt{HSMEA}.
    \item\underline{Number of interpolated solutions $\tilde{N}$}: This parameter controls the number of solutions tend to be generated by the \texttt{manifold interpolation} step and it is set as $\tilde{N}=100$ in our experiments.
    \item\underline{Number of repeated runs}: Each algorithm is independently run on each test problem for $31$ times with different random seeds.
\end{itemize}

\subsection{Performance Metric and Statistical Tests}
\label{sec:metrics}

In our experiments, we use the HV as the performance metric to assess the performance of different peer algorithms. To have a statistical interpretation of the significance of comparison results, we use the following three statistical measures in our empirical study.
\begin{itemize}
	\item\underline{Wilcoxon signed-rank test}~\cite{Wilcoxon1945IndividualCB}: This is a non-parametric statistical test that makes little assumption about the underlying distribution of the data and has been recommended in many empirical studies in the EA community~\cite{DerracGMH11}. In particular, the significance level is set to $p=0.05$ in our experiments.
    \item\underline{Scott-Knott test}~\cite{MittasA13}: Instead of merely comparing the raw HV values, we apply the Scott-Knott test to rank the performance of different peer techniques over 31 runs on each test problem. In a nutshell, the Scott-Knott test uses a statistical test and effect size to divide the performance of peer algorithms into several clusters. In particular, the performance of peer algorithms within the same cluster are statistically equivalent. Note that the clustering process terminates until no split can be made. Finally, each cluster can be assigned a rank according to the mean HV values achieved by the peer algorithms within the cluster. In particular, since a greater HV value is preferred, the smaller the rank is, the better performance of the technique achieves.
    \item\underline{$A_{12}$ effect size}~\cite{VarghaD00}: To ensure the resulted differences are not generated from a trivial effect, we apply $A_{12}$ as the effect size measure to evaluate the probability that one algorithm is better than another. Specifically, given a pair of peer algorithms, $A_{12}=0.5$ means they are \textit{equivalent}. $A_{12}>0.5$ denotes that one is better for more than 50\% of the times. $0.56\leq A_{12}<0.64$ indicates a \textit{small} effect size while $0.64 \leq A_{12} < 0.71$ and $A_{12} \geq 0.71$ mean a \textit{medium} and a \textit{large} effect size, respectively. 
\end{itemize}
Note that both Wilcoxon signed-rank test and $A_{12}$ effect size are also used in the Scott-Knott test for generating clusters.

% !tex root = main.tex

\section{Empirical Studies}
\label{sec:experiments} 

We seek to answer the following research questions (RQs) through our empirical study in the following paragraphs.
\begin{itemize}
\item\textit{\underline{RQ1}}: How is the performance comparison among our proposed six algorithm instances?
\item\textit{\underline{RQ2}}: How is the performance of our best, medium and worst algorithm instances compared against state-of-the-art algorithms in the literature?
\item\textit{\underline{RQ3}}: What is the benefit of manifold interpolation?
\item\textit{\underline{RQ4}}: What are the impacts of hyperparameters?
\end{itemize}

\subsection{Comparisons among our proposed six algorithm instances}
\label{sec:rq1}

The statistical comparison results of HV values, based on the Wilcoxon signed-rank test, among six algorithm instances introduced in~\pref{sec:instances} are given in Tables 1 to 4 of our supplementary materials\footnote{The supplementary materials can be found in~\url{https://tinyurl.com/258xne5d}.}. From these results, we can see that the HV values obtained by different algorithms are close to each other; while the best algorithm alternates across different test problem instances. 

\begin{figure}[t!]
    \centering
    \includegraphics[width=.6\linewidth]{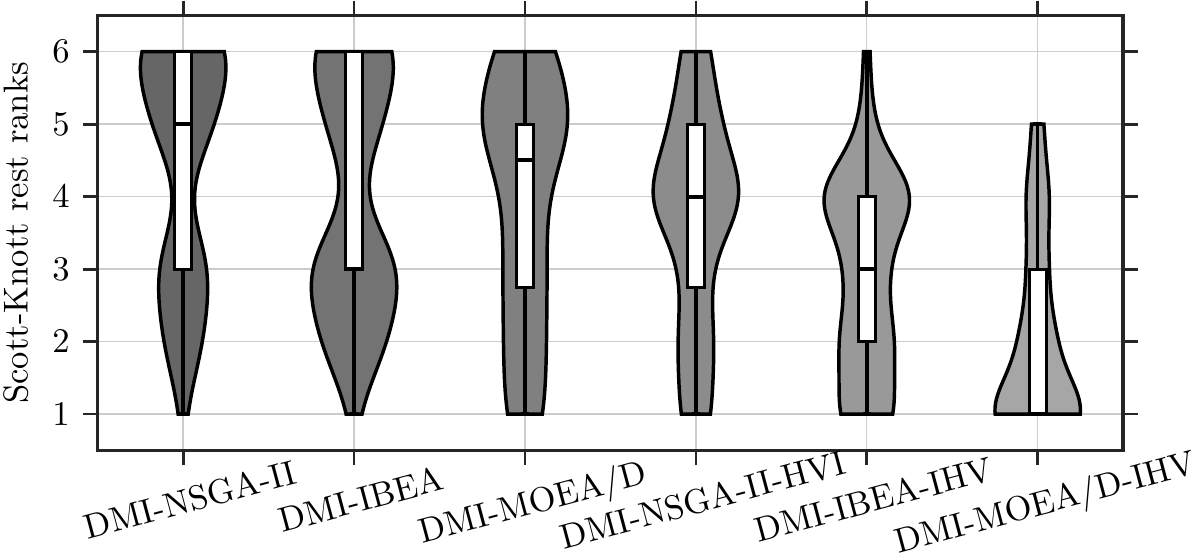}
    \caption{Violin plots of Scott-Knott test ranks achieved by each of the six algorithm instances of our proposed framework (the smaller rank is, the better performance achieved).}
    \label{fig:violin_rq1}
\end{figure}

\begin{figure}[t!]
    \centering
    \includegraphics[width=.6\linewidth]{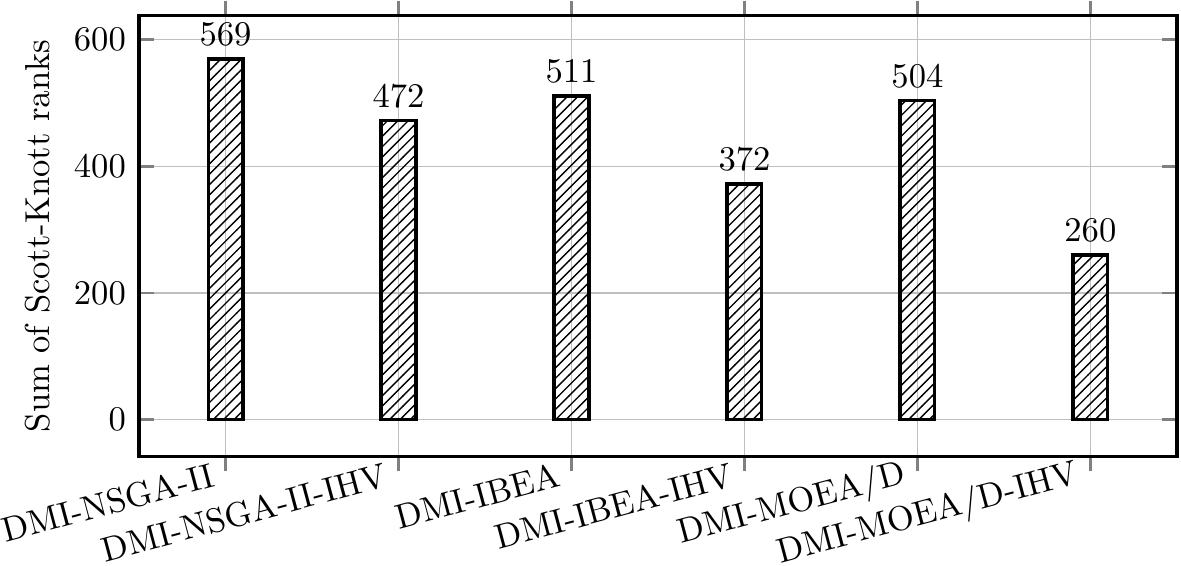}
    \caption{Total Scott-Knott test ranks achieved by each of the six algorithm instances of our proposed framework (the smaller rank is, the better performance achieved).}
    \label{fig:ranks_rq1}
\end{figure}

\begin{figure}[htbp]
    \centering
    \includegraphics[width=.4\linewidth]{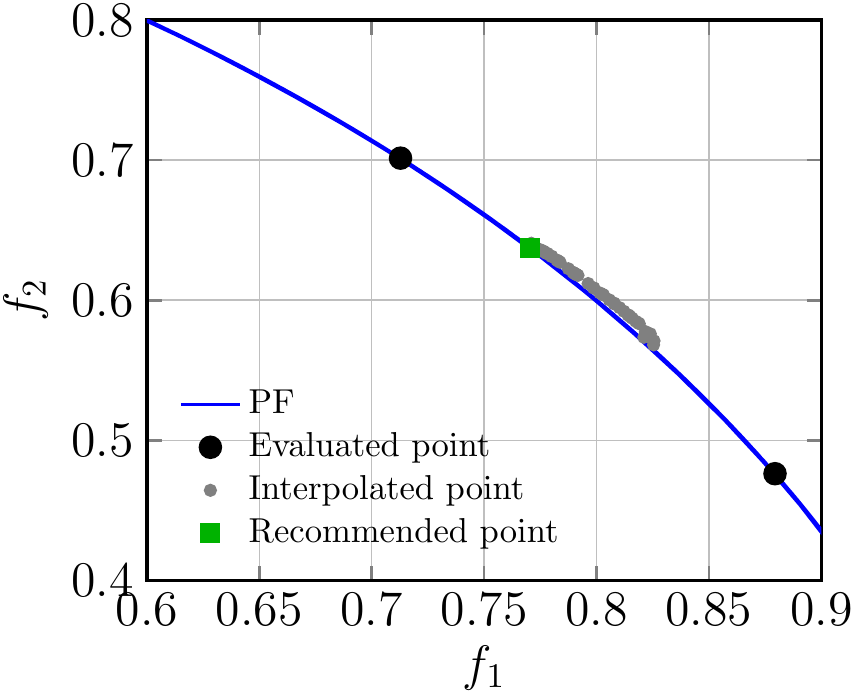}
    \caption{Illustrative example of the drawback of using the crowding distance in \texttt{DMI-NSGA-II}.}
    \label{fig:cd_example}
\end{figure}

To facilitate a better ranking among these algorithms, we apply the Scott-Knott test to classify them into different groups according to their performance on each test problem instance. Due to the large number of test problem instances used in our experiments, it will be messy if we list all ranking results ($136*6=816$ in total)  obtained by the Scott-Knott test collectively. Instead, to have a better interpretation of the comparison among different algorithm instances, we pull all the Scott-Knott test results together and show their distribution and variance as violin plots in~\pref{fig:violin_rq1}. In addition, to facilitate an overall comparison, we further summarize the Scott-Knott test results obtained across all test problem instances for each algorithm instance and show them as the bar charts in~\pref{fig:ranks_rq1}. From these results, we can see that using the IHV in the batch recommendation has shown to be consistently better than the native environmental selection mechanism in NSGA-II, IBEA and MOEA/D. In particular, we clearly see that \texttt{DMI-MOEA/D-IHV} is the best algorithm instance of our proposed framework given that 1) its performance has been classified into the best group in most comparisons as the violin plots shown in~\pref{fig:violin_rq1}; and 2) it obtains the smallest summation rank as shown in~\pref{fig:ranks_rq1} (it is at least $30$\% better than the other five peer algorithms). \texttt{DMI-NSGA-II} is the worst algorithm instance, the inferior results obtained by which can be attributed to the use of the crowding distance. In particular, due to a large number of candidates solutions generated by the manifold interpolation, the overly crowded local niche makes the crowding distance less discriminative. As the example shown in~\pref{fig:cd_example}, since the interpolated solutions are heavily crowded, the crowding distance always recommends the one lying in the boundary of the interpolated region whereas the internal solutions are ignored. In this case, it compromises the extra diversity provided by the \texttt{manifold recommendation} step. However, by using the IHV as an alternative of the crowding distance in the batch recommendation, the performance of \texttt{DMI-NSGA-II-IHV} is significantly promoted while it even obtains a better ranking than \texttt{DMI-MOEA/D} and \texttt{DMI-IBEA}.
%whereas using the IHV in the batch recommendation improves the performance. The inferior results obtained by \texttt{DMI-NSGA-II} can be attributed to the use of the crowding distance which is less discriminative when comparing a large number of solutions within a local niche. In terms of the total ranks achieved over all test problem instances, as shown in~\pref{fig:ranks_rq1}, it is interesting to note that all algorithm instances based on the IHV as the criterion in the batch recommendation generally outperform those using the native environmental selection mechanisms. In particular, \texttt{DMI-MOEA/D-IHV} is at least $30$\% better than the other five peer algorithms and even \texttt{DMI-NSGA-II-IHV} obtains a better total rank than \texttt{DMI-MOEA/D} and \texttt{DMI-IBEA}.
%\texttt{DMI-IBEA-IHV} and \texttt{DMI-IBEA} are comparable with each other. This can be explained as the fitness function used in \texttt{IBEA} is in principle similar to the IHV.

\begin{figure}[t!]
    \centering
    \includegraphics[width=.6\linewidth]{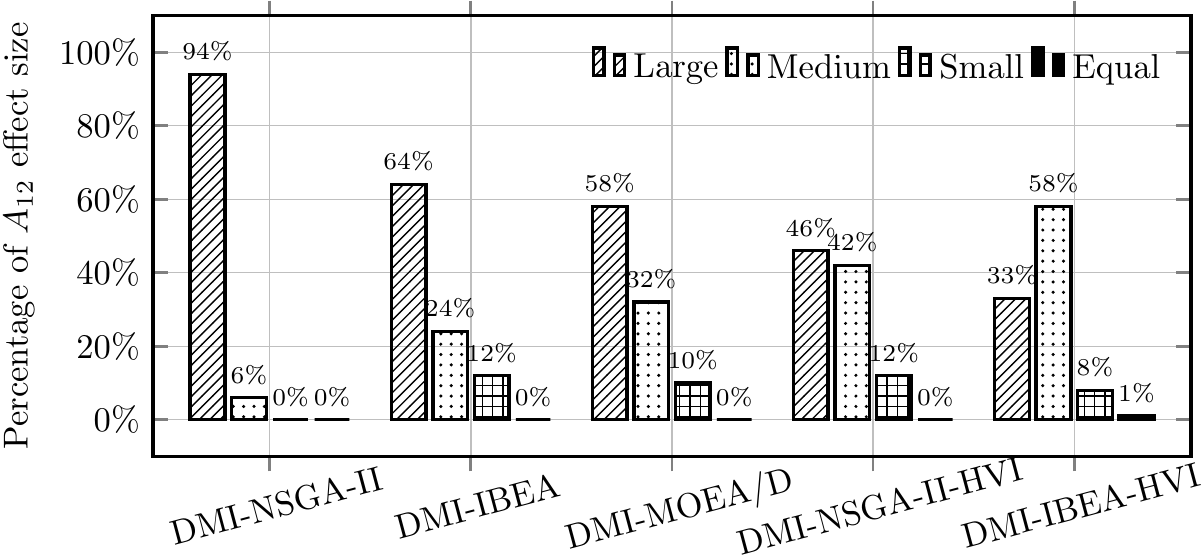}
    \caption{Percentage of the large, medium, small, and equal $A_{12}$ effect size, respectively, when comparing \texttt{DMI-MOEA/D-IHV} with other five peer algorithm instances.}
    \label{fig:a12_rq1}
\end{figure}

At the end, we choose \texttt{DMI-MOEA/D-IHV} as a representative algorithm to compare the difference of its performance with respect to the other five peer algorithms by using the $A_{12}$ effect size separately. Since the calculation of $A_{12}$ effect size is conducted in a pairwise manner, there are $136\times 5=680$ piecemeal $A_{12}$ comparison results. We again pull all results together and calculate the percentage of the equivalent, small, medium and large effect size, respectively, with respect each of the other five peer algorithms (note that there is barely equivalent case in these comparisons). From the statistical results shown in~\pref{fig:a12_rq1}, it is interesting to note that \texttt{DMI-MOEA/D-IHV} has shown dominantly better results comparing to \texttt{DMI-NSGA-II} and \texttt{DMI-NSGA-II-IHV} where the large effect sizes are all over $90$\%. In contrast, the effect sizes with regard to \texttt{DMI-IBEA}, \texttt{DMI-IBEA-IHV} and \texttt{DMI-MOEA/D} are relatively comparable. %In particular, there is a small effect size observed when comparing with its native counterpart \texttt{DMI-MOEA/D}.

\begin{tcolorbox}[breakable, title after break=, height fixed for = none, colback = gray!40!white, boxrule = 0pt, sharpish corners, top = 0pt, bottom = 0pt, left = 2pt, right = 2pt]
    \underline{Answers to \textit{RQ}1}: We have the following takeaways from our experiments. 1) \texttt{DMI-MOEA/D-IHV} is the best algorithm instance of our proposed framework while \texttt{DMI-NSGA-II-IHV} and \texttt{DMI-NSGA-II} are the medium and worst ones respectively. 2) Owing to the unique characteristics of HV for measuring convergence and diversity simultaneously, the IHV has shown to be a better mechanism for guiding the batch recommendation. 3) In contrast, the crowding distance used in NSGA-II is too coarse-grained to pick up representative solutions from a large amount of candidates; 4) \texttt{MOEA/D} is the best baseline surrogate optimizer in the \texttt{evolutionary search} step while \texttt{NSGA-II} is the worst both for using the IHV and the native environmental selection in the batch recommendation.
\end{tcolorbox}

\subsection{Comparisons with other six state-of-the-art peer algorithms}
\label{sec:rq2}

Similar to~\pref{sec:rq1}, we first pull all statistical comparison results of HV values, based on the Wilcoxon signed-rank test, among each of our six algorithm instances as introduced in~\pref{sec:instances} with regard to the other six state-of-the-art peer algorithms as introduced in~\pref{sec:peers} in Tables 1 to 4 of our supplementary materials. From these results, we find that the HV values obtained by our algorithm instances are better than the other six peer algorithms in most comparisons, even for \texttt{DMI-NSGA-II}, our least competitive algorithm instance. To have a better visual interpretation of the superiority achieved by our algorithm instances, let us look into the population distribution of the non-dominated solutions against the other six peer algorithms. Due to the page limit, we only show a couple of examples in Figs.~\ref{fig:pop_zdt31} to~\ref{fig:pop_wfg48} while the complete results can be found in the supplementary materials. From these plots, it is clear to see that our proposed algorithms not only converge well to the PF, but are also resilient to the PF shapes. Especially for those with disconnected PF segments, our proposed algorithms can approximate all segments with a reasonable diversity. In contrast, the other peer algorithms are either struggling on converging to the PF or hardly approximate all disconnected PF segments. It is interesting to note that all algorithms have shown comparable results on WFG41 to WFG48 problems with two objectives. But the performance of the other six peer algorithms degrade significantly when they go to the three-objective cases. Another interesting observation is that the increase of the number of variables do not downgrade the performance of our proposed algorithms.

\begin{figure*}[t!]
    \centering
    \includegraphics[width=\linewidth]{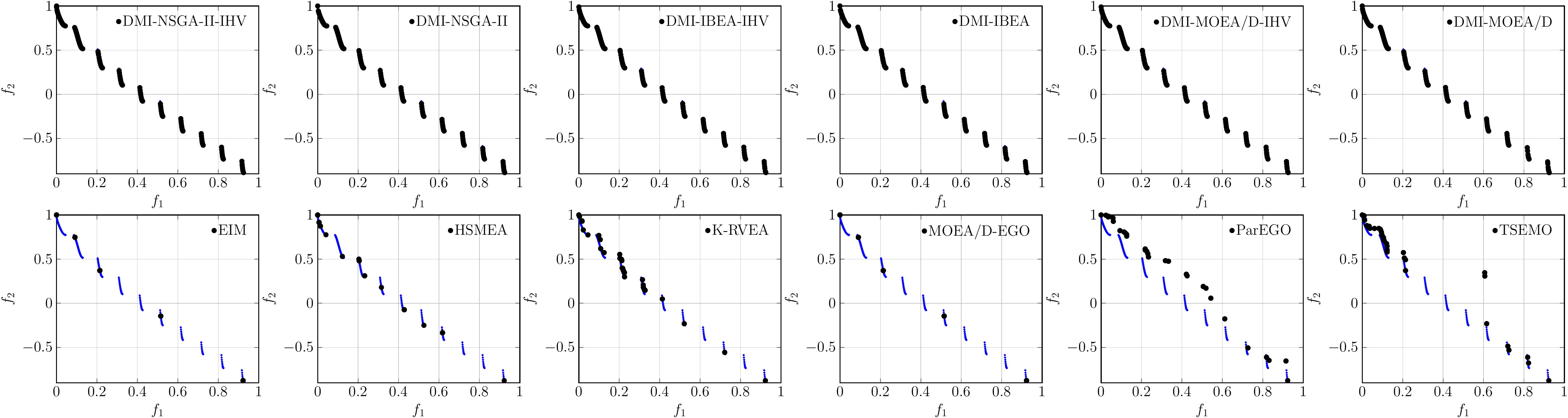}
    \caption{Non-dominated solutions found by different algorithms with the best HV values on ZDT31 ($n=30$).}
    \label{fig:pop_zdt31}
\end{figure*}

\begin{figure*}[t!]
    \centering
    \includegraphics[width=\linewidth]{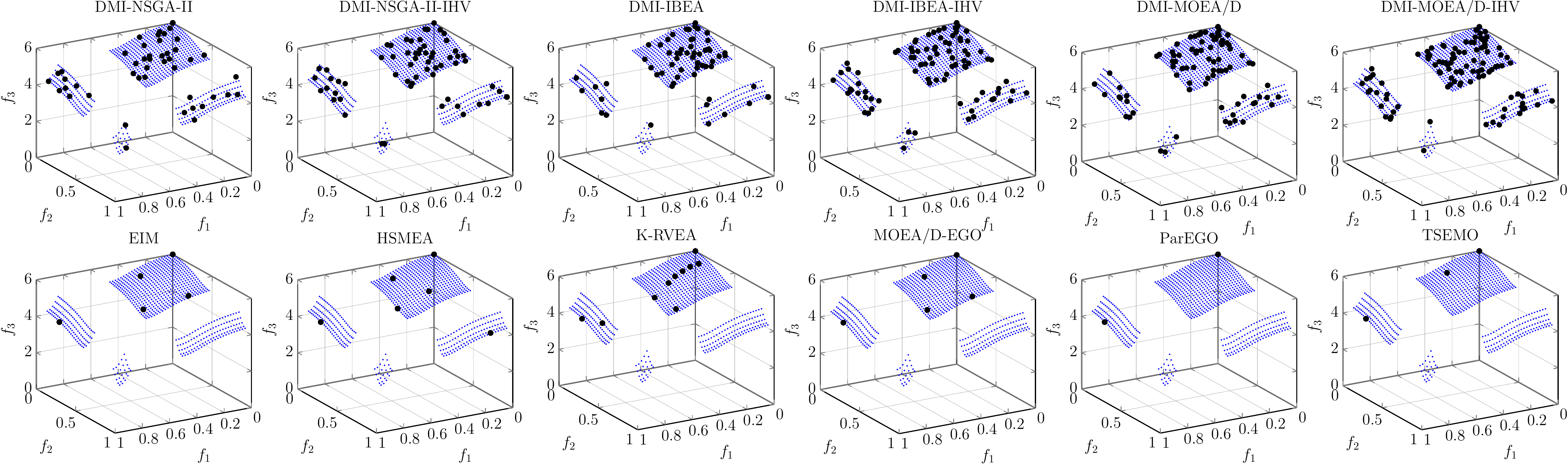}
    \caption{Non-dominated solutions found by different algorithms with the best HV values on DTLZ72 ($n=30$).}
    \label{fig:pop_dtlz72}
\end{figure*}

\begin{figure*}[t!]
    \centering
    \includegraphics[width=\linewidth]{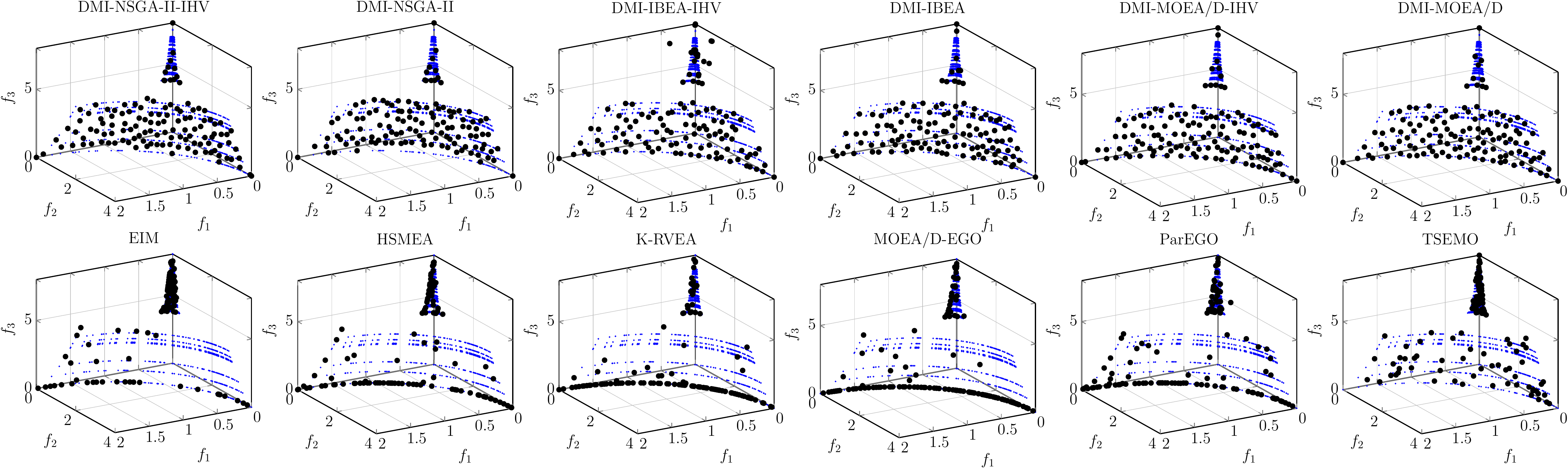}
    \caption{Non-dominated solutions found by different algorithms with the best HV values on WFG48 ($n=30$).}
    \label{fig:pop_wfg48}
\end{figure*}

\begin{figure*}[t!]
    \centering
    \includegraphics[width=\linewidth]{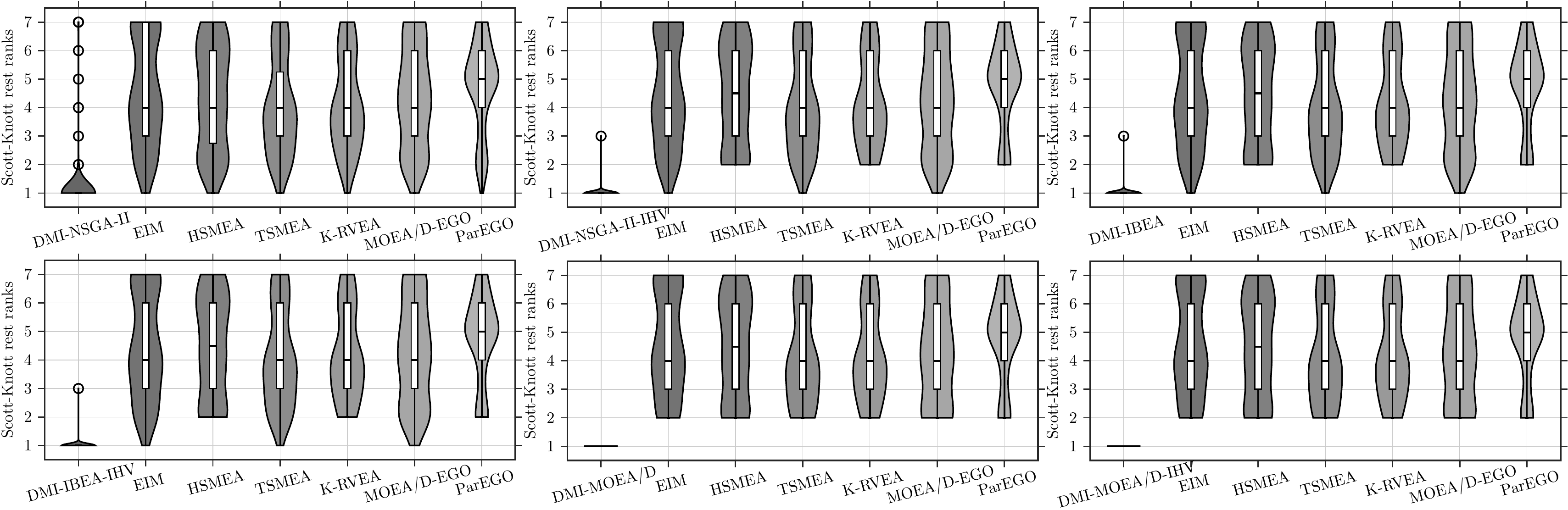}
    \caption{Violin plots of Scott-Knott test ranks achieved by each of the six algorithm instances of our proposed framework versus the other six state-of-the-art peer algorithms (the smaller rank is, the better performance achieved).}
    \label{fig:scott_knott_rq2}
\end{figure*}

As in~\pref{sec:rq1}, we apply the Scott-Knott test to sort the performance of each algorithm instance against the other six peer algorithms on all test problem instances. To facilitate a better interpretation of these massive comparison results, for each of our six algorithm instance, we pull $136\times 7\times 6=5,712$ comparison results collected from the Scott-Knott test together and show their distribution and variance as the violin plots in~\pref{fig:scott_knott_rq2}. From these results, we further confirm that our six algorithm instances are always better than the other peer algorithms in the corresponding comparisons. Specifically, \texttt{DMI-MOEA/D} and \texttt{DMI-MOEA/D-IHV} are consistently ranked in the first place in all comparisons with regard to the other six peer algorithms. In contrast, the other four algorithm instances only have very few cases that are not ranked in the best place, even for \texttt{DMI-NSGA-II}.

\begin{figure*}[t!]
    \centering
    \includegraphics[width=\linewidth]{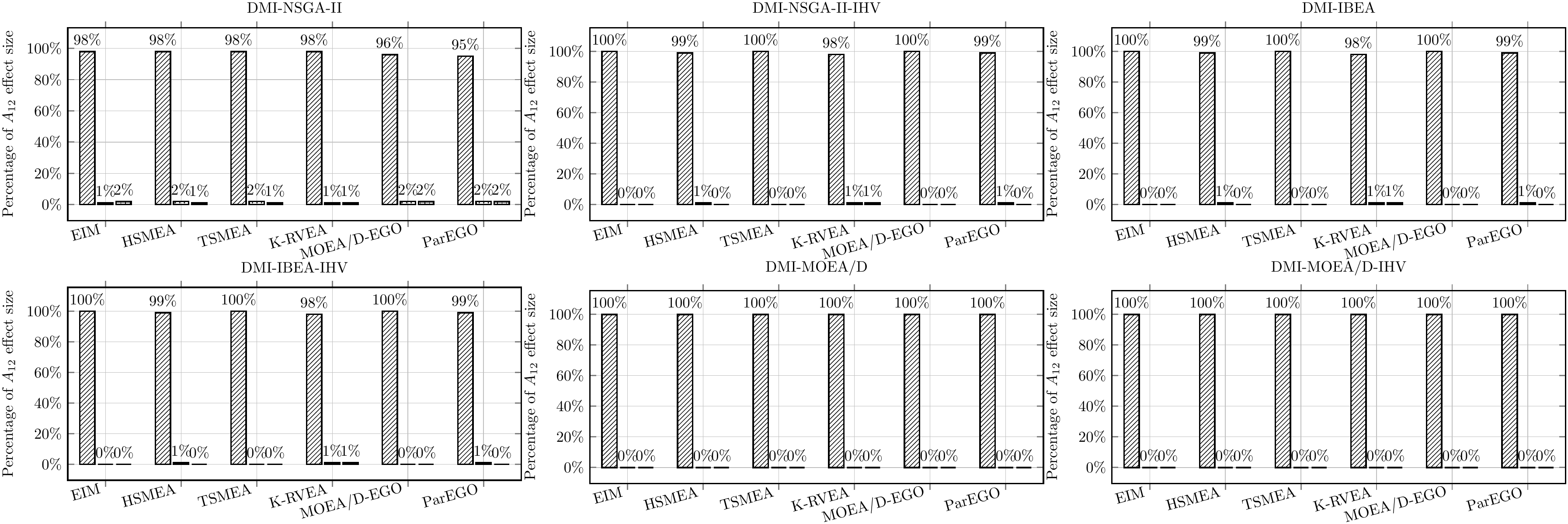}
    \caption{Percentage of the large, medium, and small $A_{12}$ effect size, respectively, when comparing each of our proposed six algorithm instances against other six state-of-the-art peer algorithms.}
    \label{fig:a12_rq2}
\end{figure*}

Again, we evaluate the $A_{12}$ effect size between each of our six algorithm instances with regard to the other six state-of-the-art peer algorithms on each test problem instance. As in~\pref{sec:rq1}, we calculate the percentage of different effect sizes obtained by each algorithm instance against the other peer algorithms, respectively. Note that since there are very few equivalent cases, we only present the results of large, medium and small $A_{12}$ effect sizes. As the bar charts shown in~\pref{fig:a12_rq2}, we further confirm the overwhelming advantage observed from the Scott-Knott test where the percentage of the large effect size is always close to $100$\% in all comparisons.

\begin{tcolorbox}[breakable, title after break=, height fixed for = none, colback = gray!40!white, boxrule = 0pt, sharpish corners, top = 0pt, bottom = 0pt, left = 2pt, right = 2pt]
    \underline{Answers to \textit{RQ}2}: We have the following takeaways from our experiments. 1) All algorithm instances of our proposed framework have shown consistently better performance over the state-of-the-art surrogate-assisted EMO algorithms in the literature, even for \texttt{DMI-NSGA-II}, our worst algorithm instance. 2) The overwhelmingly better performance achieved by our proposed framework can be attributed to the \texttt{manifold interpolation} step that help interpolates the approximated PS manifold thus significantly increases the population diversity for exploring disconnected regions.
\end{tcolorbox}

\subsection{Ablation study with regard to the manifold interpolation}
\label{sec:rq3}

The empirical study in~\pref{sec:rq2} has shown overwhelmingly better performance of our proposed framework against the selected state-of-the-art surrogate-assisted EMO algorithms. Referring to~\pref{fig:flowchart}, we can see the \texttt{manifold interpolation} step is the unique component of our proposed framework. To address RQ3, we plan to investigate the usefulness of this \texttt{manifold interpolation} step through an ablation study. To this end, we compare the performance between the algorithm instances under our proposed batched data-driven EMO framework against the  corresponding ablated counterpart without using the \texttt{manifold interpolation} step. Accordingly, it is denoted as the one without the \texttt{DMI} prefix.

\begin{figure}[t!]
    \centering
    \includegraphics[width=.6\linewidth]{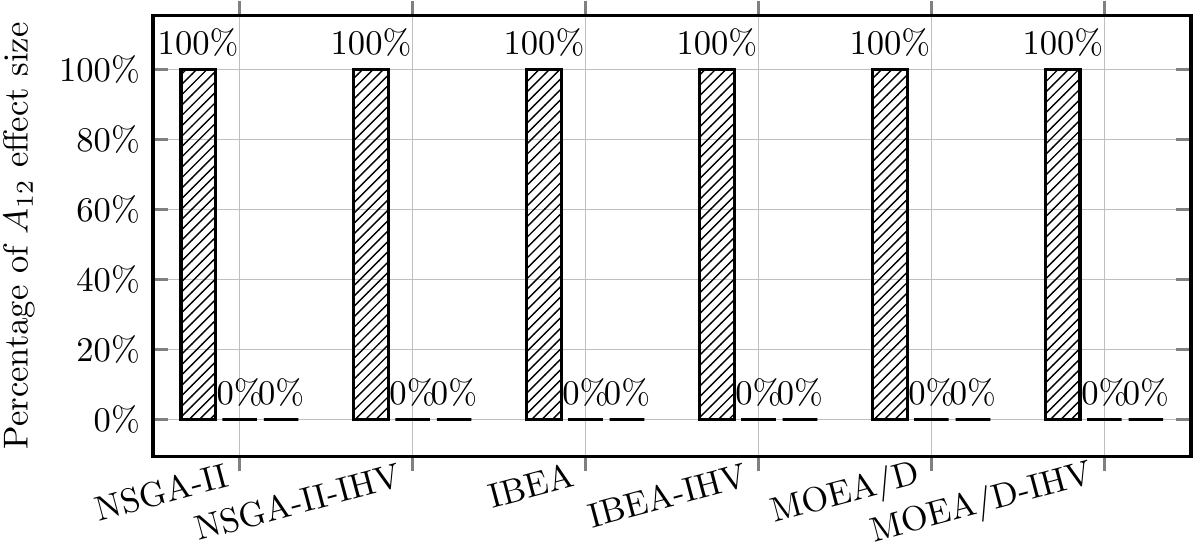}
    \caption{Percentage of the large, medium, and small $A_{12}$ effect size, respectively, when comparing each of our proposed six algorithm instances against its ablated variant without using the \texttt{manifold interpolation} step.}
    \label{fig:a12_rq3}
\end{figure}

\begin{figure}[t!]
    \centering
    \includegraphics[width=.4\linewidth]{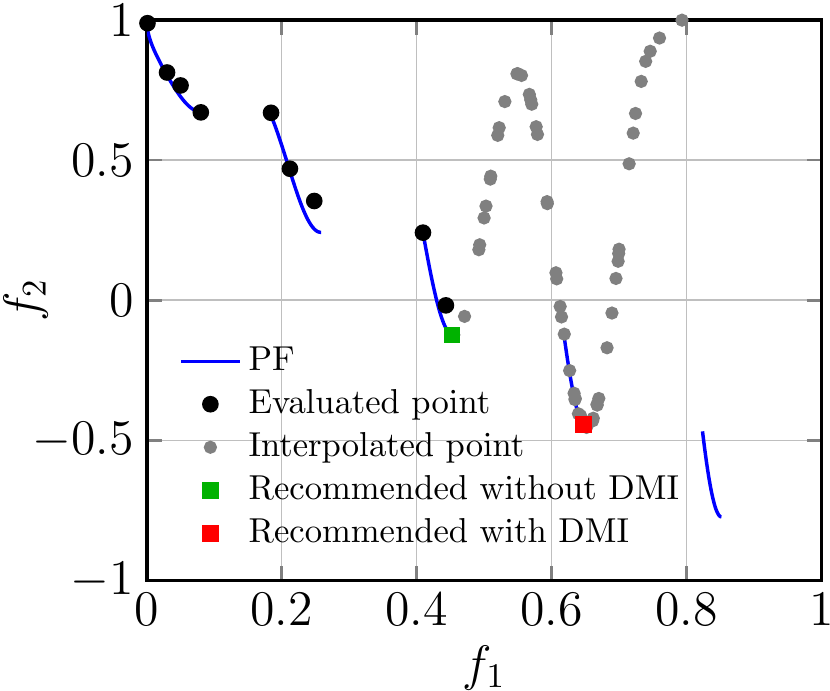}
    \caption{Illustrative example of the effectiveness of having \texttt{manifold interpolation} step.}
    \label{fig:dmi_example}
\end{figure}

From the statistical comparison results of HV values, based on the Wilcoxon signed-rank test, shown in Tables 5 and 6 in the supplementary materials along with the $A_{12}$ effect size shown in~\pref{fig:a12_rq3}, we have witnessed a clear performance degradation when ablating the \texttt{manifold interpolation} step without any exception. It is worth noting that their performance is worse than most of the selected state-of-the-art algorithms considered in~\pref{sec:rq2} by referencing Tables 1 to 4 in the supplementary materials. As an example shown in~\pref{fig:sample_rq3}, we can see that non-dominated solutions obtained by \texttt{MOEA/D-IHV} cannot fully approximate all disconnected PF segments. Without using the \texttt{manifold interpolation} step, \texttt{MOEA/D-IHV} is merely guided by the surrogate model which is highly likely to be guided to some local regions. This can be explained as the evolutionary population is far away from the PF at the early stage of the evolution. In contrast, the \texttt{manifold interpolation} step brings more diversified candidates in the survival competition. Let us consider an illustrative example shown in~\pref{fig:dmi_example}. Without using the manifold interpolation, \texttt{MOEA/D-IHV} can only obtain the solution, denoted as the green square, lying the same PF segment of previously evaluated solutions. On the other hand, because of the interpolated solutions, \texttt{DMI-MOEA/D-IHV} is able to explore under discovered PF segment as spotted by the red square. Moreover, since the interpolated solutions are along the currently approximated PF rather than purely random solutions, they are prone to have a promising convergence property.

\begin{figure}[htbp]
    \centering
    \includegraphics[width=.7\linewidth]{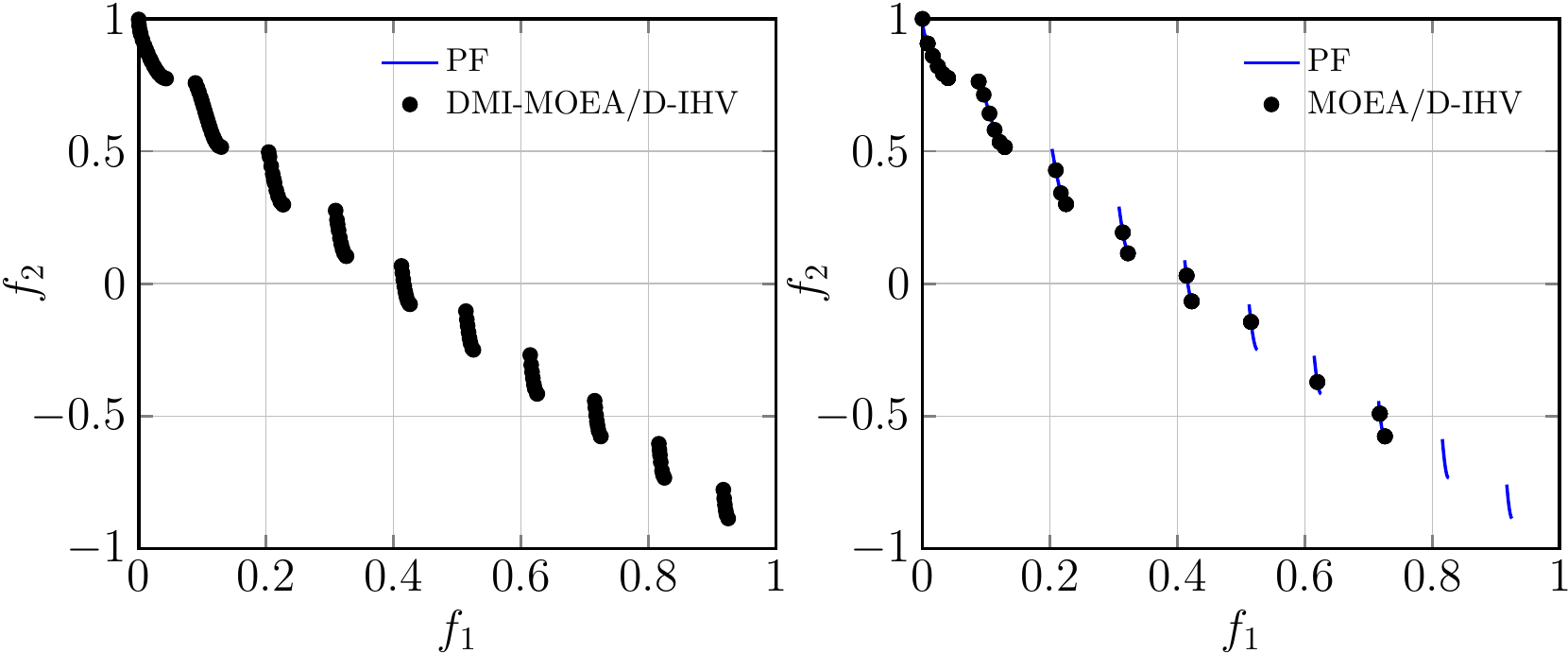}
    \caption{Comparative example of \texttt{DMI-MOEA/D-IHV} against its counterpart where the manifold interpolation is ablated on ZDT31 ($n=30$).}
    \label{fig:sample_rq3}
\end{figure}

\begin{tcolorbox}[breakable, title after break=, height fixed for = none, colback = gray!40!white, boxrule = 0pt, sharpish corners, top = 0pt, bottom = 0pt, left = 2pt, right = 2pt]
    \underline{Answers to \textit{RQ}3}: The \texttt{manifold interpolation} step is essential in our proposed framework. It not only brings sufficient diversity to expand the population, the interpolated solutions also have a promising convergence property given that they are interpolated along the approximated PS manifold. As a result, it enables our algorithms to have a faster convergence rate and a better ability to approximate different disconnected PF segments.
\end{tcolorbox}

\subsection{Parameter sensitivity study}
\label{sec:rq4}

In our proposed batched data-driven EMO framework, there are two hyper-parameters including the batch size $\xi$ and the number of interpolated solutions $\tilde{N}$ in the \texttt{manifold interpretation} step. To address RQ4, we choose \texttt{DMI-MOEA/D-IHV} as the baseline and empirically investigate its performance under different $\xi=\{5,10,20\}$ and $\tilde{N}=\{50,100,200\}$ settings.

\begin{figure}[t!]
    \centering
    \includegraphics[width=.6\linewidth]{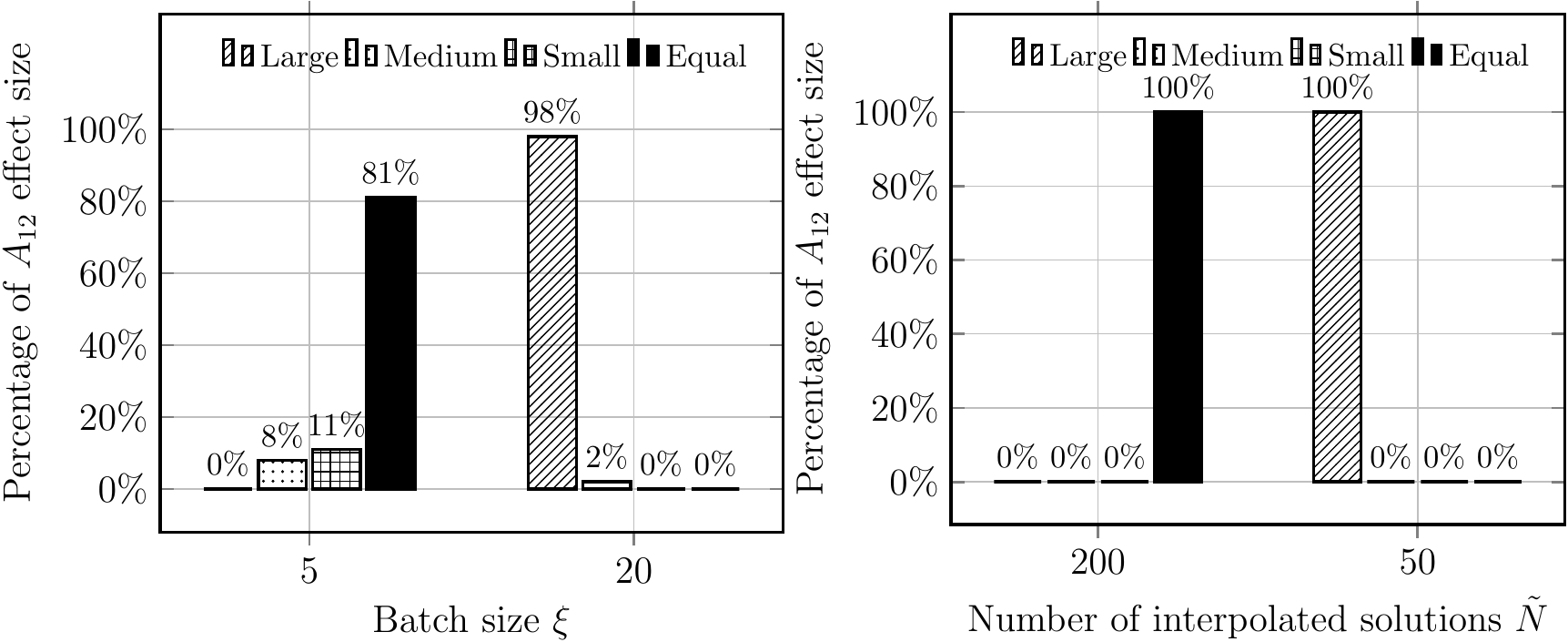}
    \caption{Percentage of the large, medium, small, and equal $A_{12}$ effect size, respectively, when comparing \texttt{DMI-MOEA/D-IHV} with our recommended $\xi$ and $\tilde{N}$ settings against others.}
    \label{fig:rq4_a12}
\end{figure}

\begin{figure}[t!]
    \centering
    \includegraphics[width=.6\linewidth]{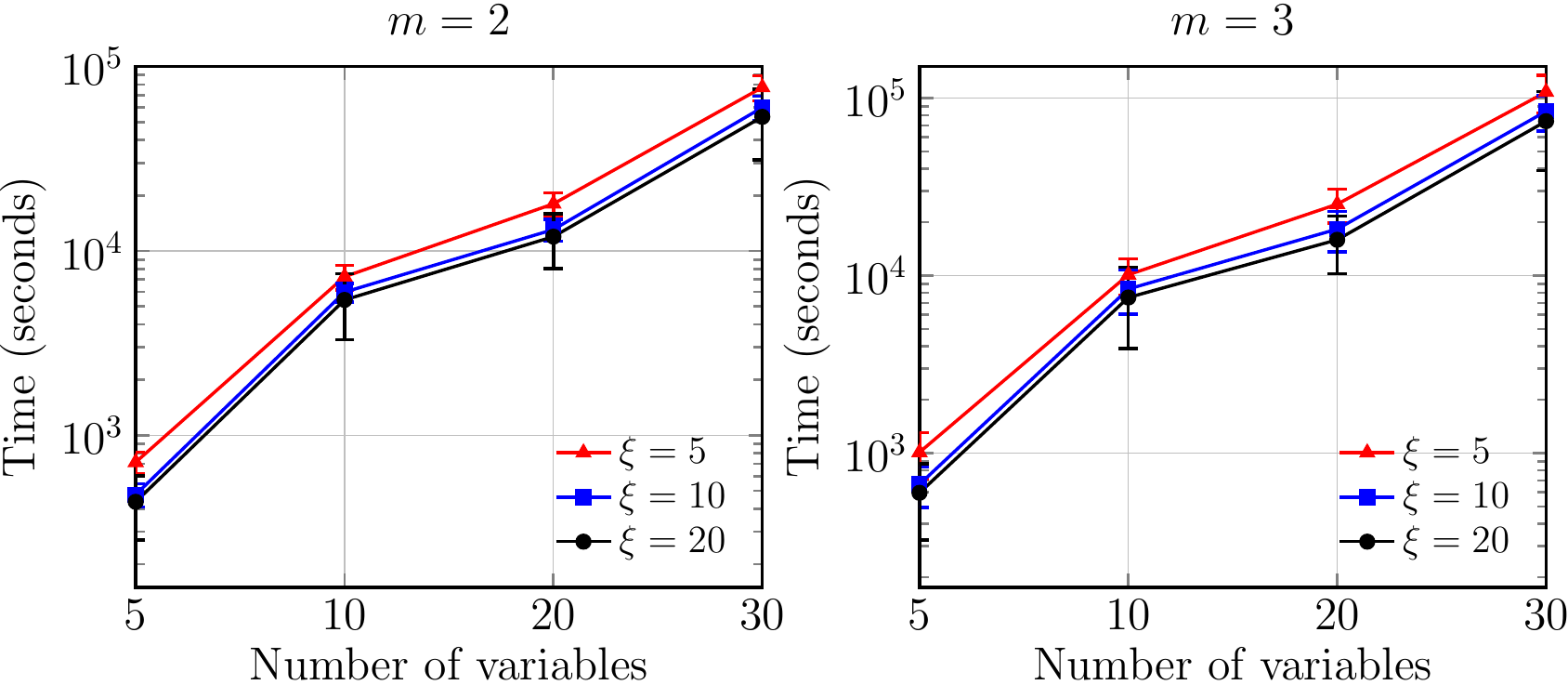}
    \caption{Collected comparisons of CPU wall clock time when using different $\xi$ settings.}
    \label{fig:time_batch}
\end{figure}

\begin{figure}[t!]
    \centering
    \includegraphics[width=.4\linewidth]{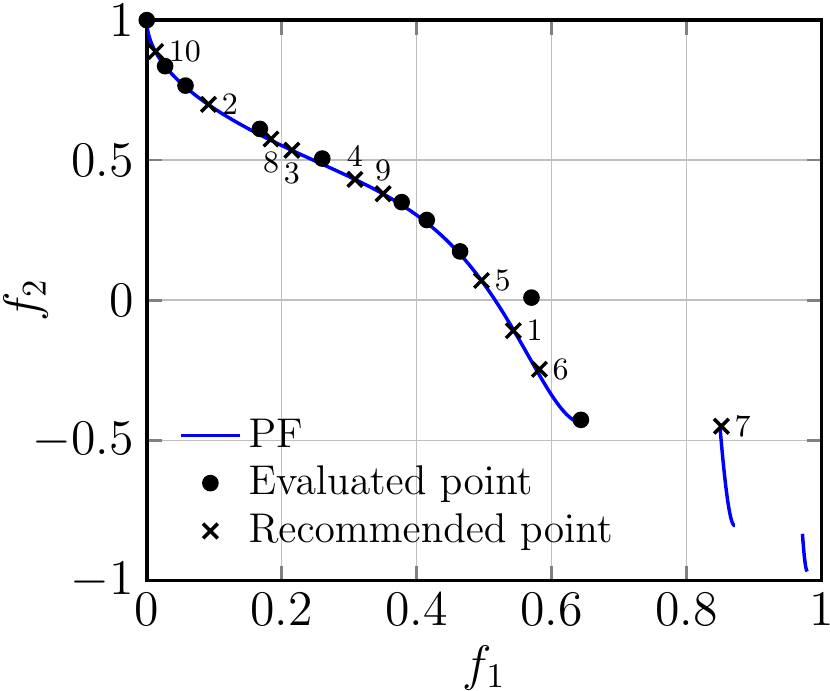}
    \caption{Illustrative example of batch recommendation results between $\xi=5$ and $\xi=10$. Specifically, Solutions \#$1$ to \#$5$ (denoted as $\times$) are recommended when $\xi=5$ while solutions \#$1$ to \#$10$ are recommended by setting $\xi=10$.}
    \label{fig:batch_example}
\end{figure}

From the statistical comparison results of HV values, based on the Wilcoxon signed-rank test, shown in Tables 7 and 8 in the supplementary materials along with the $A_{12}$ effect size shown in~\pref{fig:rq4_a12}, we can see that the performance of \texttt{DMI-MOEA/D-IHV} is the comparable when setting $\xi=5$ and $\xi=10$ where $81$\% of the comparison results are statistically equivalent. Given a limited amount of FEs, a smaller $\xi$ leads to more iterations as in our proposed framework thus it is more time consuming. \pref{fig:time_batch} gives the comparison of CPU wall clock time among different $\xi$ settings. From this figure, we can see that \texttt{DMI-MOEA/D-IHV} is multiple times slower when $\xi=5$ than those of $\xi=10$ and $\xi=20$. In addition, as an example shown in~\pref{fig:batch_example}, using a too small $\xi$ may compromise the chance for exploring under discovered PF segment(s) as solution \#$7$ (denoted as $\times$) lying in a new segment. On the other hand, although it is faster when picking up more solutions in the \texttt{bath recommendation} step by setting a larger $\xi$, the surrogate model becomes less resilient to local optima due to the reduced iterations for updating the surrogate model. As the comparison results of $A_{12}$ effect size shown in~\pref{fig:rq4_a12}, it is clear to see the large performance degradation when increasing $\xi$ to $10$.

As the comparison results shown in Tables 9 and 10 in the supplementary materials along with the $A_{12}$ effect size shown in~\pref{fig:rq4_a12}, it is interesting to note that the performance of \texttt{DMI-MOEA/D-IHV} is significantly degraded when having too small interpolated solutions (i.e., $\tilde{N}=50$) whereas it does not make statistically meaningful difference when we further increase $\tilde{N}$. However, the computational time is significantly increased in the \texttt{batch recommendation} step when having a large amount of interpolated solutions.

\begin{figure}[t!]
    \centering
    \includegraphics[width=.6\linewidth]{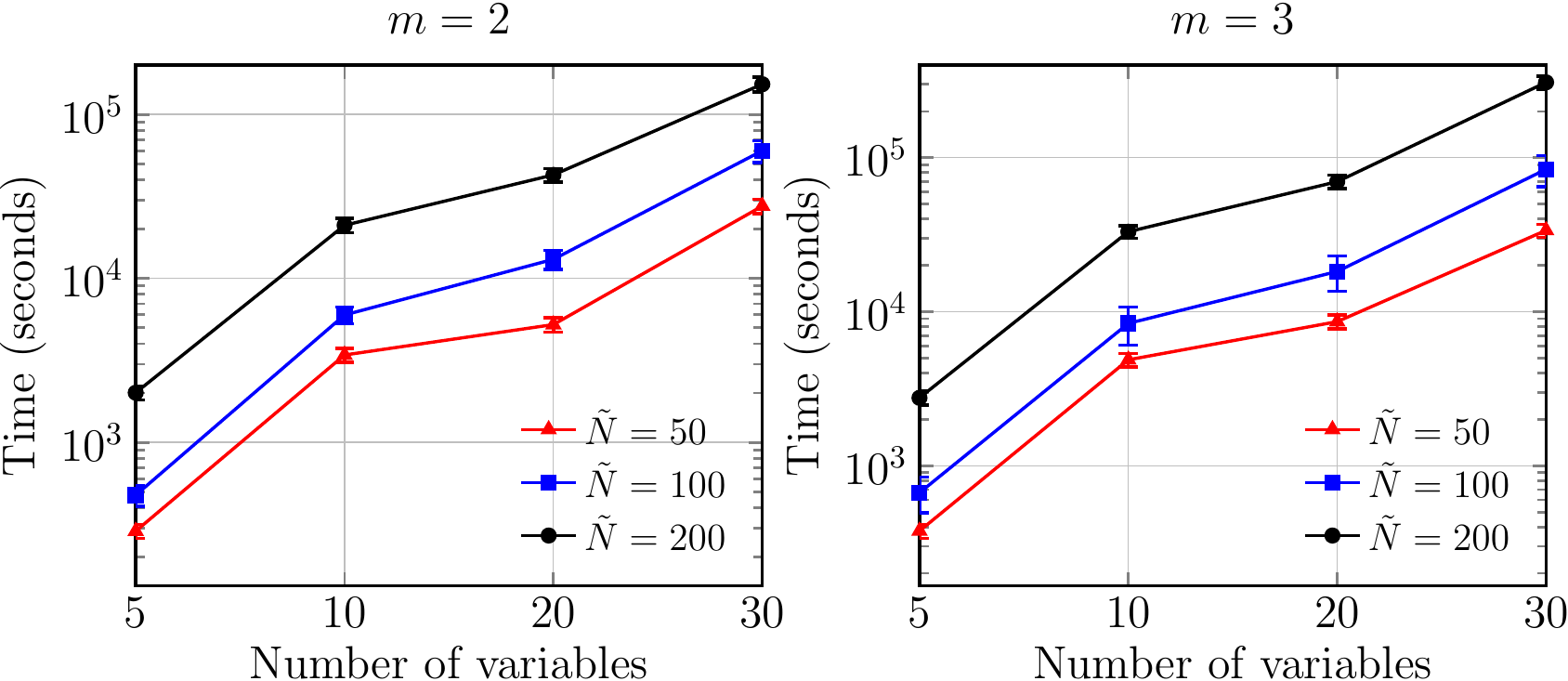}
	\caption{Collected comparisons of CPU wall clock time when using different $\tilde{N}$ settings.}
    \label{fig:time_sample}
\end{figure}

\begin{tcolorbox}[breakable, title after break=, height fixed for = none, colback = gray!40!white, boxrule = 0pt, sharpish corners, top = 0pt, bottom = 0pt, left = 2pt, right = 2pt]
    \underline{Answers to \textit{RQ}4}: We have the following takeaways from our experiments. 1) The batch size $\xi$ can influence the performance of \texttt{DMI-MOEA/D-IHV} where a too small $\xi$ makes the chosen solutions to be less representative with regard to the PF whereas a too large $\xi$ renders the surrogate model less resilient to local optima. 2) The performance of \texttt{DMI-MOEA/D-IHV} is not sensitive to the number of interpolated solutions $\tilde{N}$ generated in the \texttt{manifold interpolation} step. However, the computational time significantly soar with the increase of $\tilde{N}$.
\end{tcolorbox}

% !tex root = main.tex

\section{Conclusions and Future Directions}
\label{sec:conclusions}

This paper proposed a batched data-driven EMO framework for solving computationally expensive MOPs. It has three distinctive features. First, this framework is so general that any existing EMO algorithm can be applied in a plug-in manner as the surrogate optimizer in the \texttt{evolutionary search} step. Second, based on the KKT conditions, its \texttt{manifold interpolation} step interpolates along the approximated PS manifold to generate more diversified candidate solutions with a convergence guarantee. Last but not the least, it provides two types of approach in the \texttt{batch recommendation} step to evaluate multiple promising solutions for expensive FEs in parallel. Extensive experiments on $136$ benchmark test problem instances with various irregular PFs fully demonstrate the effectiveness and overwhelming superiority against six state-of-the-art EMO algorithms. In particular, our ablation study validates that the \texttt{manifold interpolation} step is essential within our proposed framework.

Data-driven evolutionary optimization has been an emerging area given the pressing requirements of sample-efficient real-world applications in various disciplines. In view of the strong performance and simple architecture of our proposed batched data-driven EMO framework, we envisage several aspects for future endeavors as follows.
\begin{itemize}
\item This paper only considers problems with two- and three- objectives given the already overwhelming superiority against the selected state-of-the-art. One of the future directions is to extend it for many-objective optimization problems. A typical challenge is the ineffectiveness of the sampling strategy suggested in~\pref{eq:interpolation} for high-dimensional problems. On the other hand, sampling too many candidate solutions during the \texttt{manifold interpolation} step incurs significantly mounting complexity in the \texttt{batch recommendation} step. 
\item In addition to the scalability in the objective space, the increase of the number of variables, as known as large-scale multi-objective optimization, also brings in significant challenges in both surrogate modeling and evolutionary optimization. One tentative way to combat the curse-of-dimensionality is divide-and-conquer that decomposes the original large-scale problem into smaller ones.
\item Real-world problems are usually accompanied with various constraints, the existing of which render the search space to be teared up into fragments. These lead to challenges in sampling and surrogate modeling since infeasible solutions tend to be useless in model building.
%\item Multi-modal problems are always challenging in optimization. In the context of data-driven optimization, multi-modality brings significant challenges to the surrogate modeling which hardly captures the characteristics of the underlying landscape. This further aggravates the follow-up evolutionary optimization based on the misleading surrogate model.
\item Last but not the least, many exciting real-world applications, ranging from engineering design to machine learning, are featured with multiple conflicting objectives and computationally expensive FEs. It is impactful to apply data-driven evolutionary optimization for those complex black-box problems.
\end{itemize}

\section*{Acknowledgment}
K. Li was supported by UKRI Future Leaders Fellowship (MR/S017062/1), EPSRC (2404317) and Amazon Research Awards.

\bibliographystyle{IEEEtran}
\bibliography{IEEEabrv,saemo}

\end{document}